%% file: main.tex
\documentclass{article}

\PassOptionsToPackage{numbers, compress}{natbib}

\usepackage[final]{neurips_2022}

\usepackage[utf8]{inputenc} 
\usepackage[T1]{fontenc}    
\usepackage{hyperref}       
\usepackage{url}            
\usepackage{booktabs}       
\usepackage{amsfonts}       
\usepackage{nicefrac}       
\usepackage{microtype}      
\usepackage[table]{xcolor}         
\usepackage{amsmath}
\usepackage{multirow}
\usepackage{subfig}
\usepackage{tikz}
\usepackage{color}
\usepackage{pifont}
\usepackage{amssymb}

\usepackage{caption} 
\usepackage{wrapfig}
\usepackage{enumitem}

\usepackage{xspace}
\usepackage{amssymb}
\usepackage{pifont}
\usepackage{wrapfig}

\usepackage{graphicx}
\newcommand{\eg}{\textit{e.g.}}
\newcommand{\ie}{\textit{i.e.}}
\newcommand{\cmark}{\ding{51}}
\newcommand{\xmark}{\ding{55}}

\newcommand{\etal}{\emph{et al.}}
\newcommand{\revision}[1]{\textcolor{black}{#1}}

\usepackage{enumitem}
\usepackage{tabulary}
\usepackage{tabulary,multirow,overpic,xcolor}
\newcolumntype{x}[1]{>{\centering\arraybackslash}p{#1pt}}

\newlength\savewidth
\newcommand{\tablestyle}[2]{\setlength{\tabcolsep}{#1}\renewcommand{\arraystretch}{#2}\centering\footnotesize}

\title{ST-Adapter: Parameter-Efficient Image-to-Video Transfer Learning}

\author{%
    Junting Pan$^{1}$\thanks{Equal contribution} , Ziyi Lin$^{1*}$, Xiatian Zhu$^{2}$, Jing Shao$^{1}$, Hongsheng Li$^{1,3}$\\
    $^1$Multimedia Laboratory, The Chinese University of Hong Kong~~~ \\
	$^2$Surrey Institute for People-Centred Artificial Intelligence, CVSSP, University of Surrey \\
 	$^3$Centre for Perceptual and Interactive Intelligence Limited
}

\begin{document}

\maketitle

\begin{abstract}
Capitalizing on large pre-trained models for various downstream tasks of interest have recently emerged with promising performance.
Due to the ever-growing model size,
the standard full fine-tuning based task adaptation strategy becomes prohibitively costly in terms of model training and storage.
This has led to a new research direction in parameter-efficient transfer learning.
However, existing attempts typically focus on downstream  tasks from the same modality (\eg, image understanding) of the pre-trained model.
This creates a limit because in some specific modalities, (\eg, video understanding) such a strong pre-trained model with sufficient knowledge is less or not available.
In this work, we investigate such a novel cross-modality transfer learning setting,
namely {\em parameter-efficient image-to-video transfer learning}.
To solve this problem, we propose 
a new {\bf \em Spatio-Temporal Adapter} (ST-Adapter) for parameter-efficient fine-tuning per video task.
With a built-in spatio-temporal reasoning capability in a compact design, 
ST-Adapter enables a pre-trained image model without temporal knowledge to reason about dynamic video content at a small ($\sim$8\%) per-task parameter cost, requiring approximately 20 times fewer updated parameters compared to previous work.
Extensive experiments on video action recognition tasks show that our ST-Adapter can match or even outperform the strong full fine-tuning strategy and state-of-the-art video models, whilst enjoying the advantage of parameter efficiency. Code and model are available at
\url{https://github.com/linziyi96/st-adapter}
\end{abstract}

\input{01-Introduction}

\input{02-RelatedWork}

\input{03-Method}
\input{04-Experiments}
\input{05-Conclusion}

\clearpage
{
\small
\bibliographystyle{plainnat}
\bibliography{egbib}
}

\clearpage
\input{checklist}
\clearpage
\input{06-Appendix}
\end{document}

%% file: 01-Introduction.tex
\section{Introduction}

In the NLP field, almost all the state-of-arts across a wide range of downstream tasks have been achieved by adapting from large pretrained models (a.k.a. {\em foundation models}~\cite{foundation}) such as BERT \cite{bert} and GPT \cite{gpt2,gpt3}. 
The {\em de facto} standard approach to adapting a pretrained model to down-stream tasks is {\em fine-tuning} either {\em fully} or {\em partially} (\eg, linear probing by training the newly added multi-layer perceptron layers on the top alone),
subject to the condition of adopting a similar network architecture as the pretrained model.
Nonetheless, given increasingly larger whilst ever stronger foundation models (\eg, GPT-3 with 175B parameters), fully fine-tuning the whole model for every single downstream task would become prohibitively expensive and infeasible in terms of training cost and model storage. 
This could significantly restrict their deployment and usability in real-world applications.
In this context, a series of NLP works has been introduced towards efficient transfer learning with better trade-offs between parameter and accuracy  \cite{adapter,unified_adapter,prefix, lester-etal-2021-power}.

This trend has recently motivated the computer vision community.
For example, the CLIP model \cite{clip}, trained with 400 million web image-text pairs, achieves promising performances on a variety of image recognition and generation tasks.
In the video domain, with significantly more computational cost and resources, Xu \etal~\cite{xu2021videoclip} trained a video variant of CLIP but excelled on a smaller number of downstream video tasks.
This is partly attributed to two orders of magnitude more minor training data and limited availability of computing resources, as large video data is notoriously more difficult to collect, manage, and process than image data.
Under these restrictions, large pre-trained image models are arguably still favorable in the selection of model initialization for video tasks.

\begin{figure}[t]
    \centering
    \includegraphics[width=1\linewidth]{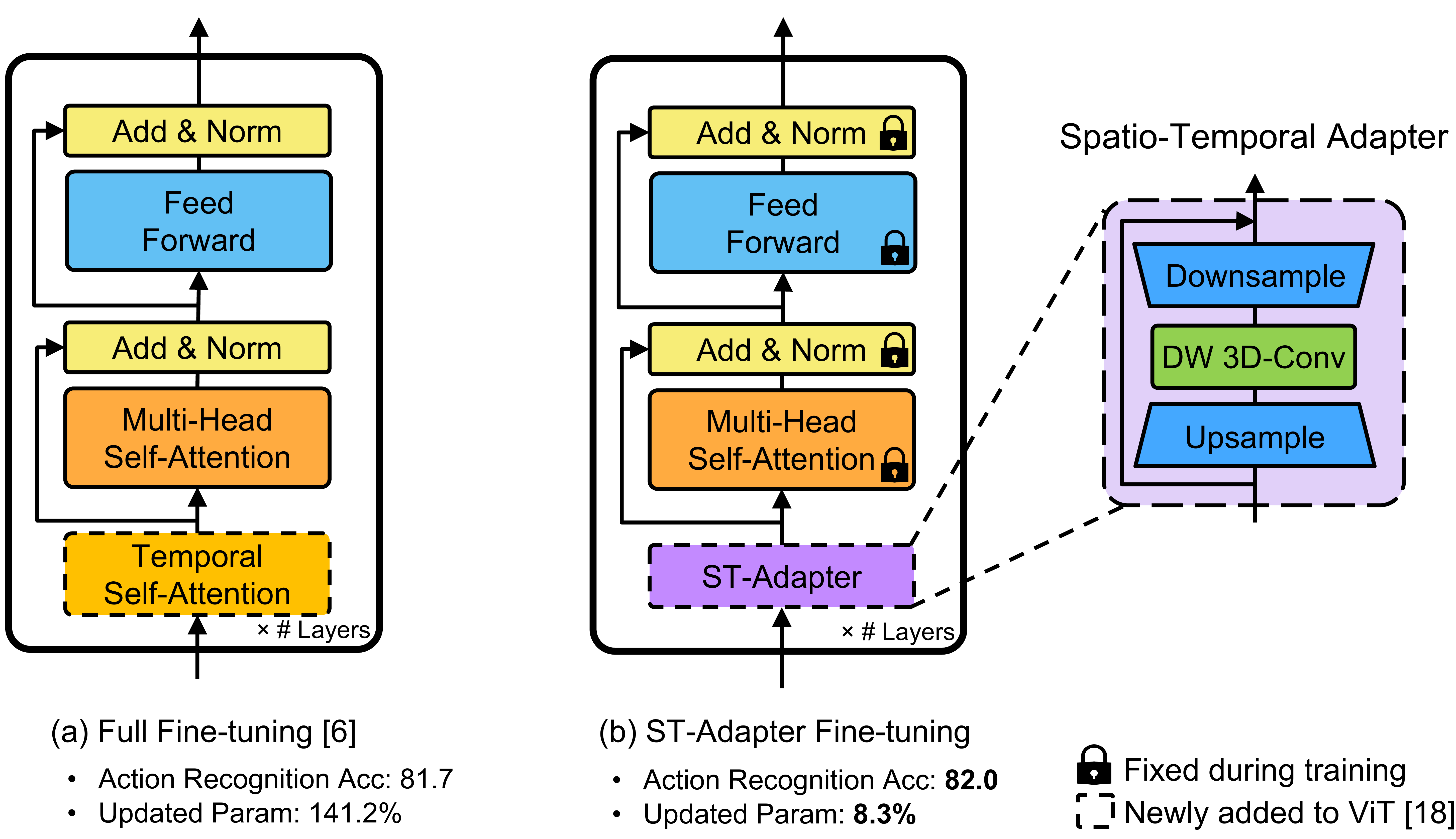}
    \caption{ \textbf{Image-to-video transfer learning strategies.
    } 
    (a) The state-of-the-art methods for adapting a pre-trained image model (\eg, ViT \cite{vit} in this example) to video tasks (\eg, action recognition)
    usually adopt the paradigm of first designing a temporal learning module and then fine-tuning the whole network fully \cite{vivit,timesformer,xvit}.
    This is parameter-inefficient since a specific instance of such a large model 
    is resulted for each downstream task.
    In contrast, (b) we propose to only train a lightweight {\em Spatio-Temporal Adapter} with 
    much fewer parameters for each individual downstream task at a significantly smaller computational cost.
    Surprisingly, our method can match or even surpasses the full fine-tuning based methods (including prior art video models in terms of accuracy),
    whist enjoying higher parameter efficiency and cheaper training cost. 
    }
    \label{fig:fig1}
    \vspace{-5mm}
\end{figure}

In this work, we investigate a novel, critical problem of {\bf\em efficiently adapting large pre-trained image models for video downstream tasks}, 
with a focus on the widely influential action recognition task.
Considering that training video models is drastically more expensive in both computing resource and time than image models \cite{feichtenhofer2019slowfast},
this problem becomes particularly more useful and valuable in practice.
On the other hand, it is also more challenging and non-trivial due to the extra necessity of overcoming the big gap between image and video in transfer learning.
Especially, pre-trained image models lack the ability to infer temporal structured information, which however is critical in video understanding.
In fact, the key design with state-of-the-art video models \cite{i3d,tsm,timesformer,xvit} is usually about learning the temporal dimension based on 
contemporary image models. 
Although model initialization is still important, they largely go beyond the fine-tuning strategy, as 
{\em architectural modification} is often imposed in addition to full model training/fine-tuning per downstream task.

Given that this is a new problem, we first conduct a comprehensive benchmark using both various fine-tuning methods for image-to-video transfer learning and state-of-the-art video models \cite{timesformer,xvit}.
Regarding the pretrained image model, we select two Vision Transformer (ViT) \cite{vit} models, with one from CLIP pre-training \cite{clip}
and the other pre-trained on ImageNet-21K \cite{imagenet_cvpr09}.
ViT is representative in terms of network architecture, pre-training algorithm, and training data scale.
Crucially, we further propose an efficient yet effective {\bf\em Space-Time Adapter} (ST-Adapter), capable of extracting and leveraging the pre-trained knowledge of a large image model to achieve superior video understanding at a small parameter cost.
Specifically, ST-Adapter is formulated based on a novel parameter-efficient bottleneck
with a sequence of operations including feature dimension reduction, spatial-temporal modeling, and feature dimension recovery.
It is easy to implement and scalable for deployment since all the primitive steps are realized with standard operators (\eg, fully-connected layer, depth-wise 3D convolution).
With such a lightweight design, our bottleneck can be cheaply integrated throughout the base network for enabling stronger layer-wise spatio-temporal learning.
As a result, our model can be more rapidly optimized using fewer training epochs for significant convergence advantage.

We summarize the {\bf contributions} as follows.
{\bf(1)} We investigate a new problem of {\em parameter-efficient image-to-video transfer learning}. Our motivation is to advocate the usability and deployment of increasingly larger whilst ever more powerful {\em pre-trained} image models in benefiting more challenging video understanding tasks.   
{\bf(2)} We establish a benchmark for action recognition tasks by comprehensively experimenting with a variety of fine-tuning strategies and several state-of-the-art video understanding models.
{\bf(3)} We introduce a novel parameter-efficient {\em Spatio-Temporal Adapter} (ST-Adapter) for more effectively capitalizing a large pre-trained image model in video understanding. 
By grounding all the primitives on standard operators, 
ST-Adaptor is easy to implement and friendly to deployment.
{\bf(4)} 
Extensive experiments on action recognition datasets show that our ST-Adapter outperforms not only existing parameter-efficient alternatives and the full fine-tuning strategy,
but also state-of-the-art video methods with the same network architecture and model initialization.

%% file: 02-RelatedWork.tex
\section{Related Work}

{\flushleft \bf Parameter-efficient transfer learning }
Driven by the wider application of large pre-trained language models across a diversity of downstream tasks, the topic of efficient tuning has received increasing attention in NLP. 
Existing efficient tuning methods fall broadly into three categories.
The {\em first} category is to introduce task-specific adapters \cite{adapter,unified_adapter,pfeiffer2020AdapterHub,pfeiffer2020adapterfusion}.  
Specifically, an adapter consists of lightweight modules inserted between layers of a pre-trained model. 
To be parameter-efficient, only those newly added adapter modules need 
to be updated during task fine-tuning, whilst all the parameters of the large pre-trained model, which takes the majority proportion of the whole solution, are frozen.
The {\em second} category is prompt tuning \cite{prefix,qin2021learning,vidprompt,shin2020autoprompt,liu2021p}.
Instead of manipulating the network architecture, 
these methods prepend a set of learnable tokens at the input point of the model or intermediate layers.
Similarly, only these added tokens need to be optimized for each downstream task. 
The {\em third} category is learning weight approximation \cite{lora}.
In particular, only the low-rank matrices for approximating the weights
need to be updated during training.

Early works for efficient transfer learning in vision focus on parameter sharing in the context of multitask learning \cite{sidetune, efficientparam, residualadapter}. Recently, there are several works for extending the efficient tuning idea from NLP to vision tasks. CoOp \cite{coop} and CoCoOp \cite{cocoop} apply prefix tuning for adapting the CLIP model to various image recognition tasks. VL-Adapter \cite{vladapter} achieves the performance comparable to full fine-tuning on challenging vision-language tasks. 
Commonly, their design focuses are all restricted to the text encoder of the CLIP model. More recently, \cite{vpt,vp,zhang2022neural} introduce the idea of prompt learning to visual backbones. They obtained favorable results on various image recognition benchmarks. Moving a step further, 
in this work, we consider the more challenging adaptation problem from a pre-trained image model without temporal knowledge to video understanding tasks.

{\flushleft \bf Video action recognition } 
Action recognition in the unconstrained video has largely been dominated by deep learning methods, thanks to the availability of large video datasets, \eg, Kinetics \cite{i3d,carreira2018short,carreira2019short} and Something-Something \cite{goyal2017something}.
As a key component, the model architectures adopted by existing video methods
has expanded from CNNs \cite{karpathy2014large,tran2015learning,feichtenhofer2019slowfast,feichtenhofer2020x3d,s3d,tran2018closer, wang2018temporal, pan2021, tsm, liu2020tam} to Transformers \cite{fan2021multiscale,li2021improved,li2022uniformer,liu2021video,vivit, timesformer}.
As temporal information is important for modeling the dynamics,
a variety of motion learning techniques has been introduced \cite{wang2018non,jiang2019stm,mformer}.
Further, different training methods have also been explored, \eg, unsupervised learning \cite{tong2022videomae,feichtenhofer2021large,wei2021masked}, and video-text contrastive learning \cite{sun2019videobert,xu2021videoclip,xu2021vlm,tian2020contrastive}. 
New opportunities for stronger video models are created following the introduction of 
large pretrained foundation models \cite{clip,jia2021scaling,yu2022coca}. For example, Wang \etal{} \cite{wang2021actionclip} equipped the CLIP with temporal modules and good performance can be achieved after the model is fully fine-tuned on video datasets. Ju \etal{} \cite{vidprompt} adopted the CLIP model for video recognition tasks by learning video-specific prompts. 
In contrast, 
in this work, we explore the potential of the large pre-trained image models
with the parameter-efficient adapter strategy.
Importantly, despite the simplicity, we bring about more significant 
advantages in performance along with a new benchmark on parameter-efficient image-to-video transfer learning.

%% file: 03-Method.tex
\section{Methodology}
To capitalize a large pre-trained image model for more challenging video understanding such as action recognition in a cross-modality manner, 
it is necessary to fill the intrinsic gap between image and video.
For easier understanding, we start with an intuitive baseline based on temporal aggregation. 

{\flushleft \bf Temporal aggregation } 
A straightforward baseline method of exploiting a pre-trained image model for video understanding is 
to temporally aggregate per-frame feature representations
(\eg, average pooling).
Concretely, given an input video clip $\mathbf{V} \in \mathbb{R}^{T \times H \times W}$, where $T, H, W$ are the number of frames, height and width respectively. Following \cite{vit}, we first split each frame into $N = H\times W / P^2$ patches of size $P \times P$. Then, we flatten these patches and project them into a sequence of patch tokens $\mathbf{Z}_{t} = [\mathbf{z}_1, ...\mathbf{z}_s,...,\mathbf{z}_N], \mathbf{z}_s \in \mathbb{R}^{d}$ where $d=3\times P^2$ with $t=1,...,T$. The sequence of feature vectors is then enhanced with the positional embedding by element-wise addition,
along with a trainable class token concatenated. 
Subsequently, we feed each sequence with $N+1$ tokens to a stack of self-attention based blocks individually.
For each sequence we keep only the classification token $\mathbf{z}_t^{cls}$. We further perform temporal average pooling on the class tokens $\mathbf{z}_{final} = \frac{1}{T}\sum\limits_t \mathbf{z}^{cls}_{t}$  to yield a compact representation for the whole clip.
We obtain the prediction by passing $\mathbf{z}_{final}$ through a classifier. 
As the sptial
information is only naively averaged over time, 
it is also known as 
\texttt{Space-Only TimeSformer} \cite{timesformer}.

{\flushleft \bf Spatio-temporal attention }For more dedicated structural modeling in the time dimension with ViTs, a mainstream approach in the video domain is to develop various spatio-temporal attention mechanisms by further imposing temporal attention on top \cite{timesformer, vivit,ba2016layer,xvit, tokenshift, hashiguchi2022vision}.
We choose two representative video ViT models, TimeSformer \cite{timesformer} and XViT \cite{xvit}, 
in our performance benchmark.
However, state-of-the-art video ViT models often need to {\em fully fine-tuned} per task,
which is {\bf \em parameter-inefficient}, given that in this way we have to keep a separate copy of the whole fine-tuned model parameters for every single task.

\subsection{Preliminaries}

Our method is inspired by the Adapter \cite{adapter}
designed for parameter-efficient transfer learning in NLP.
Specifically, the adapter module is composed of a down-projection linear layer followed by a non-linear activation function and an up-projection linear layer. Formally, given an input feature matrix $\mathbf{X} \in \mathbb{R}^{N \times d}$ at the $i$-th layer, the feature adaptation process can be written as: 
\begin{equation} \label{eq:adapter}
    \texttt{Adapter}(\mathbf{X}) = \mathbf{X} + f(\mathbf{X}\mathbf{W}_{down})\mathbf{W}_{up},
\end{equation}
where $\mathbf{W}_{down} \in \mathbb{R}^{d \times r}$ refers to the down projection layer,  $\mathbf{W}_{up} \in \mathbb{R}^{r \times d}$ the up-projection layer, and $f(\cdot)$ the activation function. Note, that a residual summation is applied for preserving the information in input as required.
The idea of Adapter has been remarkably successful in NLP
due to several advantages: (1) High parameter efficiency across tasks since only a small number of parameters are task-specific; 
(2) Reaching on-par performance compared to full fine-tuning;
(3) Taking significantly small training costs; 
(4) Avoiding the catastrophic forgetting limitation of full fine-tuning.

We aim to propagate the success of Adapter from NLP to computer vision
particularly the image-to-video transfer learning problem as discussed earlier.
To that end, we introduce a novel Adapter tailored specially for spatio-temporal reasoning -- a key capability for video understanding which, however, existing NLP Adapter variants lack.

\subsection{Spatio-Temporal Adapter (ST-Adapter)} 

Typically, an image model only considers the ability of spatial modeling.
The {\em objective} of our Spatio-Temporal Adapter (ST-Adapter) is to enable a pre-trained image model to reason about spatial and temporal information of video in a parameter efficient principle.
In design, we consider a couple of practically-crucial criteria: 
(1) {\em Smaller parameter size}: The parameter cost for each downstream task 
should be small -- the essential criterion for parameter efficiency.
(2) 
{\em Development friendliness}: This is critical for real-world development and deployment. In practice, it is necessary that a model can be easily implemented using the standard highly optimized deep learning toolboxes (\eg, PyTorch, TensorFlow, TensorRT, and TorchScript), without tedious per-toolbox specialization.
This also facilitates the realization of high inference efficiency
across a diversity of running platforms due to the best usage of
built-in software and hardware resources.

Under these considerations, we formulate the proposed ST-Adapter by sticking to {\em commonly-adopted primitive operators alone}. 
Starting with the above Adapter (Eq. \eqref{eq:adapter}) originally developed for NLP tasks, we further introduce a spatio-temporal operator realized by a standard depth-wise 3D-convolution layer \cite{feichtenhofer2020x3d} between the bottlenecks (Figure \ref{fig:fig1}). 
In particular, our spatio-temporal operator enables layer-wise temporal inference efficiently, because it only operates in a compressed low-dimensional (\eg, 128D) feature space and the depth-wise convolution is highly efficient both in parameter and computation \cite{mobilenet}.
As a result, this yields an introduction of tiny extra ($\sim$2\%) parameters and ($\sim$0.3\%) computation.
Formally, our ST-Adapter can be expressed as:
\begin{equation}
    \texttt{ST-Adapter}(\mathbf{X}) = \mathbf{X} + f\Big(\texttt{DWConv3D}(\mathbf{X}\mathbf{W}_{down})\Big)\mathbf{W}_{up},
\end{equation}
 where $\texttt{DWConv3D}$ denotes the depth-wise 3D-convolution for spatio-temporal reasoning we introduce. It is noteworthy that before applying $\texttt{DWConv3D}$, the down-projected feature representations will be first reshaped from $\mathbf{X'} \in \mathbb{R}^{T \times N \times d}$ to  $\mathbf{X''} \in \mathbb{R}^{T \times h \times w \times d}$  (where $ N = h \times w$)
to have the spatial and temporal dimensions prepared for reasoning.
With this highly integrated design, our ST-Adapter enjoys the same efficiency and flexibility as the NLP Adapter, while uniquely being able to conduct spatio-temporal modeling. 
l.

\subsection{ST-Adapter Integration} 
For proper adaptation, the adapter modules are often integrated between layers of a Transformer.
In NLP, a variety of integrating designs have been investigated.
For example, \cite{adapter} deploys two adapter modules per layer with one following the Multi-Head Self-Attention (MHSA) and the other following the Feed-Forward Networks (FFN) \cite{adapter}.
On the other hand, \cite{stickland2019bert,bapna2019simple} suggest that adding only one adapter after the FNN suffices. 
Similarly, our ST-Adapter can be also integrated generally at distinctive positions.
Empirically, we find that a decent performance can be achieved in case a single ST-Adapter is placed before the MHSA of each transformer block (Figure \ref{fig:fig1}(a) and Table \ref{tab:ablation:local_pos}).

%% file: 04-Experiments.tex
\section{Experiments}
\label{sec:Experiments}

\subsection{Experiments Setup}

{\flushleft \bf Datasets} For the benchmark experiments, we use two popular video action recognition datasets.

\textit{Kinetics-400 (K400)}: The K400~\citep{kay2017kinetics} dataset contains $\sim$240k training videos and 20k validation videos labeled with 400 action categories.
Most videos have a length of 10s or about 300 frames.
While there is a great diversity in these videos, they are largely biased to spatial appearance \cite{sevilla2021only}.

\textit{Something-Something-v2 (SSv2)}: The SSv2~\citep{goyal2017something} dataset consists of 220,487 videos covering 174 human actions. The video length ranges from 2 to 6 seconds.
In contrast to K400, SSv2 presents richer temporal information with much higher significance \cite{sevilla2021only}.

\textit{Epic-Kitchens-100 (EK100)}: \revision{The EK100~\citep{epic} dataset consists of 100 hours of video in egocentric perspective recording a person interacting with a variety of objects in the kitchen. Each video sample is labeled with a verb and a noun. We report top-1 verb and noun classification accuracy.}

{\flushleft \bf Pre-trained models} 
In all experiments, we use the standard ViT \cite{vit} as our base backbone model. We conduct most of our experiments with the \textit{ViT-B/16} variant with 12 layers and 86M parameters, taking as input a sequence of patches at size $16\times16$.

What was learned during pre-training directly decides the knowledge that can be transferred to downstream tasks, thus also the effectiveness upper bound of transfer learning methods. To this end, we benchmark the same backbone under two different pre-training strategies: pre-training with web-scale raw data that has been recently proposed by CLIP \cite{clip} (400M image-text pair) and classical supervised pre-training on annotated data from ImageNet-21K (21k classes and 14M images).

{\flushleft \bf Implementation details.}~All details, including training and testing settings and module instantiation details, are provided in the appendix.

{\flushleft \bf Competitors } We provide several transfer learning approaches in our benchmark for efficient image-to-video transfer learning. Note that the parameters of the linear classifier are always updated during training for all approaches. 

\begin{itemize}[leftmargin=2em, topsep=0.15mm]
\item[(1)]\textit{Full Fine-tuning}: Fully updating all the parameters when adapting for a specific target task. 

\item[(2)]\textit{Partial Fine-tuning}: Only update the last ViT layer while keeping the rest of the parameter fixed.

\item[(3)]\textit{Temporal Fine-tuning}: We only tune the temporal attention modules ({\it i.e.,} TA) in the SA+TA architecture.

\item[(4)]\textit{Linear Probing}: Freezing all the parameters except those in the linear classification layer.

\item[(5)]\textit{Adapter} \cite{adapter}: \revision{Adding small sub-networks between layers of a pre-trained model. During fine-tuning, we only update the newly added parameters introduced by the adapters.}

\item[(6)]\textit{Prompt Tuning} \cite{vpt}: Prepending a sequence of learnable prompt tokens to the input visual patch tokens. During fine-tuning, only these newly added prompts are updated.

\item[(7)]\textit{Attention Pooling Head}: Replacing the original temporal average pooling with a temporal attention pooling layer (similar to the one used in \cite{xvit}) before the classification head.  
\end{itemize}

These approaches above do not incorporate temporal modeling to the image ViT.
Hence, we further consider temporally augmented ViT architectures
as introduced in state-of-the-art video methods: 
\begin{itemize}[leftmargin=2em, topsep=0.15mm]
\item[(a)]\textit{Spatial Attention Only (SA)}: Space-Only TimeSformer \cite{timesformer}.

\item[(b)]\textit{Spatial Attention + Temporal Attention (SA+TA)}: The default TimeSformer \cite{timesformer} with divided space-time attention (Fig.~\ref{fig:fig1}a).

\item[(c)]\textit{Spatial Attention + Temporal Shift (SA+TS)}: XViT \cite{xvit}.
\end{itemize}

Note that not all fine-tuning protocols are compatible with each of these video ViT variants. Take SA+TS for example, the original model behavior is altered with channel shift, as a result, it is not compatible with \textit{Linear Probing} that requires freezing all the parameters of the backbone.

\subsection{Main Results and Analysis}

{\flushleft \bf Cross-modality fine-tuning benchmark.} Table~\ref{tab:compare_efficient_transfer} presents the results of fine-tuning a ViT-B/16 pre-trained with CLIP and ImageNet-21K. All baselines are built by combining existing efficient fine-tuning methods with three state-of-the-art ViT-based action recognition models. From the results we can see that:

(i) For CLIP pre-trained model, ST-Adapter performs on par with Full Fine-tuning (\textbf{82.0} vs. 81.7 for K400 and \textbf{66.3} vs. 66.1 for SSv2) while updating far less parameters (\textbf{7.2M} vs. 121.57M).
ST-Adapter significantly outperforms all other efficient fine-tuning methods. 
We see that baselines like Prompt Tuning and Partial Fine-tuning can provide non-trivial gain in performance compared to Linear Probe, but are still behind our ST-Adapter.

(ii) \revision{Our ST-Adapter can generalize across different pre-training datasets and methods. We can see that} CLIP pre-train models dominate over ImageNet-21K pre-train ones. These results well match the shift of paradigm in current AI research~\citep{foundation}, where pre-training no longer needs limiting to curated data and annotations to deliver good performance on downstream tasks, but can take advantage of broader scale web raw data.

Interestingly, we observe that SSv2, a motion-centric dataset in design, also benefits from stronger appearance (image) pre-training. We think this may attribute to that raw textual description can provide a much richer description (\ie,~human-object relations) of the image than curated limited categorical labels.
Full fine-tuning on SA+TS (XViT) performs slightly worse with CLIP pretrain than ImageNet-21k pretrain. We conjecture this is because the channel shift operation breaks the knowledge in the pre-training weights, and thus does not benefit \revision{much} from \revision{stronger pre-training like CLIP.} 

\input{tables/compare_efficient_transfer}

{\flushleft \bf Comparison to the state-of-the-art models.} \revision{We compare ViT with ST-Adapter to other state-of-the-arts methods on both K400 dataset \citep{kay2017kinetics} in Table~\ref{tab:sota_kinetics}, SSv2 dataset \citep{goyal2017something} in Table~\ref{tab:sota_ssv2} and EK100 dataset \citep{epic} in Table~\ref{tab:epic}. }We can observe that:

(i) With the proper adaptation method, we can simply turn a large image foundation model into a good video model by only tuning a few parameters. Our results are comparable to or better than previous methods tailored for such tasks. Our largest model with ViT-L backbone set a new state-of-the-art in K400 by achieving 86.7\% top-1 accuracy.

(ii) It is noteworthy that, our method takes significantly fewer frames as input compared to other methods (8 vs. 16, 32, 64, 96). It is also reflected in terms of GFlops. Saying that the ViT was not designed for efficiency purposes like \citep{li2022uniformer, xvit,liu2021swin,fan2021multiscale} but the adapted CLIP ViT has achieved similar accuracy-efficiency trade-offs.

(iii) The paradigm of pre-training and fine-tuning has been widely adopted in most state-of-art methods to achieve good performance. Between them, most of the approaches start from image pre-trained models, and only a few can afford video pre-training. Note that for the Something-Something dataset, except MViT~\citep{fan2021multiscale} pre-trained on video data from scratch, the rest of methods are still initialized from image pre-trained weights. A good image pre-trained model with rich appearance information can facilitate temporal modeling in temporally challenging datasets like SSv2. 

\revision{(iii) It is evident in Table \ref{tab:epic} that our ST-Adapter consistently brings a big margin on egocentric videos. Also, we found that without our ST-Adapter, it is much more difficult to directly adapt CLIP pre-trained ViT on the domain of egocentric video with high sensitivity to the hyper-parameter setting. ST-Adapter eases the training process. It is worthy to note that, all current transformer based approaches need to be pre-trained first on image dataset and then fine-tuned on Kinetics dataset before fine-tuned with egocentric videos. In contrast, our ST-Adapter can be directly applied to an image model and trained with target egocentric video alone.}

\begin{table}[!t]
\centering
\caption{\textbf{Results on Kinetics-400 validation set}. “Frames” denotes the total number of frames used during inference which is:  \# frames per clip $\times$ \# temporal clip $\times$ \# spatial crop. “GFlops” means $10^9$ Flops.
Our ViT w/ ST-Adapter achieves new state-of-the-art performances on K400 at similar GFlops.
} 
\vspace{-1mm}
\tablestyle{4.8pt}{1.1}
\input{tables/sota_k400}
\label{tab:sota_kinetics}
\end{table}

\begin{table}[!t]
\centering
\caption{\textbf{Results on Something-Something-v2 validation set.} 
“Frames” denotes the total number of frames used during inference which is:  \# frames per clip $\times$ \# temporal clip $\times$ \# spatial crop. “GFlops” means $10^9$ Flops. Our ViT w/ ST-Adapter outperforms most of the current methods by only fine-tuning a very small set of parameters. Here the ViT-B w/ ST-Adapter result is reported using {\bf 2 ST-Adapters per block}.}  
\vspace{-1mm}
\tablestyle{4.8pt}{1.1}
\input{tables/sota_ssv2}
\label{tab:sota_ssv2}
\end{table}

\input{tables/sota_ek100}

\subsection{Ablations}

\input{figs/multiple_fig_in_a_row}

Unless otherwise specified, we use ViT-B/16 backbone and 8 input frames in all ablation experiments, and we use one ST-Adapter with bottleneck width 384 before MHSA in each Transformer block.

\input{tables/ablation}

{\flushleft \bf Where to insert ST-Adapter}  By default, we insert a ST-Adapter to every Transformer block in the backbone, but we also show the performance impact of using fewer ST-Adapters. As shown in Table \ref{tab:ablation:global_pos}, while more ST-Adapters tend to do better, ST-Adapters at deeper layers boost performance more than those at shallower layers. This observation is useful when we insert ST-Adapters into deeper models and having an Adapter for each block might be too expensive. We also show the performance when inserting ST-Adapters to different positions within a block. As shown in Table \ref{tab:ablation:local_pos}, while the performance is relatively insensitive to the position of the Adapters, using multiple adapters in one block may substantially boost performance on some datasets, like SSv2 in our case.

\vspace{-2mm}
{\flushleft \bf Training parameter efficiency} We experiment with a different number of channels in the middle of the bottleneck design. As shown in Table \ref{tab:ablation:channel} and Fig.~\ref{fig:num_params}, our method is effective with a wide range of bottleneck width: even with a channel reduction to 64, our ST-Adapters still obtain relatively good performance, outperforming all baselines in Table \ref{tab:compare_efficient_transfer} except for Full Fine-tuning (SA + TA). 
Even with a bottleneck width of 768, our ST-Adapters are still very parameter efficient, introducing only about 1/6 new parameters to a Transformer encoder block. In contrast to the {\em inverted bottleneck} design commonly used with depthwise convolutions \cite{sandler2018mobilenetv2}, ST-Adapters work best with regular bottlenecks.
The success of transfer learning with such low-rank projections again shows the rich knowledge and strong potential of modern foundation models.

\vspace{-2mm}
{\flushleft \bf Training time efficiency} In Fig.~\ref{fig:train_epoch} 
we show an enlarged difference between full fine-tuned models and our ST-Adapters with low training budgets. When we reduce the number of training steps, the accuracy of full fine-tuned models drops significantly faster than models with ST-Adapters. This shows the advantage of our proposed modules when backbone models are large or computational resources are limited. \revision{We also report the total training GPU-hours and peak memory usage for three models: TimeSformer, ViT-B/16, ViT-B/16 with ST-Adapter (8 input frames, 16 samples per GPU on 8 V100 GPUs) in Table \ref{tab:train_time}.}

{\flushleft \bf Training data efficiency} Fig.~\ref{fig:data_size} showcases the impact of training data size on action recognition accuracy. Even with the same pre-trained weights, ST-Adapters tend to obtain higher performance than full fine-tuning especially on smaller datasets: the margin between the two models increases with the shrinkage of data. This shows that ST-Adapters are powerful tools to transfer to downstream tasks where only a small amount of labeled data is available.

{\flushleft \bf Effects of kernel shape}\revision{~We ablate the effect of kernel size in the depth-wise convolutions inside our proposed ST-Adapter. It is shown in Table~\ref{tab:kernel_shape} that the temporal span is most sensitive, suggesting the significance of temporal structural modeling as we focus on in this work.}

\input{tables/training_time}
\input{tables/kernel_shape}

%% file: tables/compare_efficient_transfer.tex
\begin{table}[t]
    \centering
    \caption{ \textbf{Benchmark results on Kinetics-400 and Something-Something-v2.} We evaluate all the approaches on two datasets with ViT-B/16 pretrained with CLIP and ImageNet-21K. For each entry, we report the top1 action recognition accuracy and the number of fine-tuned parameters.
    All methods introduce extra parameters beside parameters 
    of the ViT backbone and linear classifier. Our ST-Adapter achieves the best trade-off between accuracy and training efficiency. It is the only efficient fine-tuning method that can match the performance of full fine-tuning. The \textit{TM?} column shows whether the method includes temporal modelling, \ie, a temporal aggregation method other than average pooling. All models are trained using 8 frames and tested with 3 views.}
    \vspace{3mm}
    \begin{tabular}{ll|cc|cc|cc}
    \toprule
     & & & & \multicolumn{2}{c|}{CLIP} & \multicolumn{2}{c}{ImageNet-21K} \\
    \cmidrule{5-8}
    \begin{tabular}{c}Fine-tuning\\Methods\end{tabular} & Architecture &
    TM? &
    \begin{tabular}{c}Fine-tuned\\Params (M)\end{tabular} & K400 & SSv2 & K400 & SSv2 \\
    \midrule
    \multirow{3}{*}{Full Fine-tuning} & SA & & 86.11  & 81.0 & 44.0 & 76.9 & 40.0 \\
    & SA + TA \cite{timesformer} & \cmark & 121.57  & \underline{81.7} & \underline{66.1} & \underline{78.0} & 59.5 \\
    & SA + TS \cite{xvit} & \cmark & 93.79 & 78.0 & 62.0 & \textbf{78.5} & \textbf{64.4} \\
    \midrule
    \multirow{2}{*}{Partial Fine-tuning} & SA & & 7.40 & 80.1 & 37.6 & 61.7 & 20.4 \\
    & SA + TA & \cmark & 10.36 & 80.3 & 57.5 & 63.1 & 29.3 \\
    \midrule
    Temporal Fine-tuning & SA + TA & \cmark & 35.8 & 81.3 & 59.4 & 76.5 & 51.9 \\
    \midrule
    Prompt Tuning & SA & & 1.18  & 79.3& 39.3 & 71.4 & 26.3\\
    \midrule
    Attentional Pooling & SA & \cmark & 2.36 & 75.3 & 21.5 & 59.1 & 15.1  \\
    \midrule
    Linear Probe & SA & & 0.31 & 76.6 & 21.9 & 60.1 & 14.8 \\
    \midrule
    \revision{Adapter \cite{adapter}} & \revision{SA} & & \revision{6.77} & \revision{81.6} & \revision{46.2} & \revision{76.2} & \revision{40.5} \\
    \midrule
    ST-Adapter (ours) & SA& \cmark & 7.20 & \textbf{82.0} & \textbf{66.3} & 76.6 & \underline{62.8}\\
    \bottomrule
    \end{tabular}
    \label{tab:compare_efficient_transfer}
    \vspace{-3mm}
\end{table}

%% file: tables/sota_k400.tex
\begin{tabular}{@{}l|x{55}|x{45}|x{45}|x{45}|x{45}@{}}
\toprule
 Model & Pretrain & \#Frames & GFlops & Top-1 & Top-5\\
\toprule
\multicolumn{6}{@{}l}{\it Methods with full-finetuning} \\
LGD\citep{lgd} & IN-1K & 128$\times$N/A & N/A & 79.4 & 94.4 \\
SlowFast+NL\citep{feichtenhofer2019slowfast} & - & 16$\times$3$\times$10 & 7020 & 79.8 & 93.9 \\
ip-CSN\citep{csn} & Sports1M & 32$\times$3$\times$10 & 3270 & 79.2 & 93.8\\
CorrNet\citep{corrnet} & Sports1M & 32$\times$3$\times$10 & 6720 & 81.0 & -  \\
X3D-XL\citep{feichtenhofer2020x3d} & - & 16$\times$3$\times$10 & 1452 & 79.1 & 93.9 \\
MoViNet-A6\citep{movienet} & - & 120$\times$1$\times$1 & 386 & 81.5 & 95.3 \\
ViT-B-VTN \citep{video_transformer} & IN21K & 250$\times$1$\times$1 & 3992 & 78.6 & 93.7 \\
TimeSformer-L\citep{timesformer} & IN21K & 96$\times$3$\times$1 & 7140 & 80.7 & 94.7  \\
STAM \citep{stam} & IN21K & 64$\times$1$\times$1 & 1040 & 79.2 & - \\
X-ViT\citep{xvit} & IN21K & 16$\times$3$\times$1 & 850 & 80.2 & 94.7 \\
Mformer-HR\citep{mformer} & IN-21K & 16$\times$3$\times$10 & 28764 & 81.1 & 95.2 \\
MViT-B,32$\times$3\citep{fan2021multiscale} & - & 32$\times$1$\times$5 & 850 & 80.2 & 94.4 \\
ViViT-L\citep{vivit} & JFT300M & 16$\times$3$\times$4 & 17352 & 82.8 & 95.3 \\
Swin-B\citep{liu2021video} & IN1K & 32$\times$3$\times$4 & 3384 & 80.6 & 94.6  \\
Swin-L(384)\citep{liu2021video} & IN21K & 32$\times$5$\times$10 & 105350 & 84.9 & 96.7 \\
UniFormer-B\citep{li2022uniformer} & IN1K & 32$\times$1$\times$4 & 1036 & 82.9 & 95.4\\
VATT-Large(320)\citep{vatt} & HowTo100M & 32$\times$3$\times$4 & 29800 & 82.1 & 95.5\\
TokenLearner\citep{tokenlearner} & JFT300M &
64$\times$3$\times$4 & 48912 & 85.4 & 96.3\\
\revision{OMNIVORE(Swin-L)\citep{omnivore}} & \revision{IN22K+SUN} &
\revision{32$\times$3$\times$4} & \revision{7248} & \revision{84.1} & \revision{96.3}\\
\revision{MTV-H\citep{mtv}} & \revision{WTS-280} &
\revision{32$\times$3$\times$4} & \revision{73570} & \revision{\textbf{89.9}} & \revision{\textbf{98.3}}\\
\revision{ViT-B w/o ST-Adapter} & \revision{CLIP} & \revision{8$\times$3$\times$1} & \revision{419} & \revision{81.0} & \revision{95.5} \\
\revision{ViT-L w/o ST-Adapter} & \revision{CLIP} & \revision{8$\times$3$\times$1} & \revision{1941} & \revision{85.8} & \revision{97.2} \\
\midrule
\multicolumn{6}{@{}l}{\it Methods with frozen backbone} \\
Our ViT-B w/ ST-Adapter & CLIP & 8$\times$3$\times$1 & 455 & 82.0 & 95.7 \\
Our ViT-B w/ ST-Adapter & CLIP & 16$\times$3$\times$1 & 911 & 82.5 & 96.0 \\
Our ViT-B w/ ST-Adapter & CLIP & 32$\times$3$\times$1 & 1821 & 82.7 & 96.2 \\
Our ViT-L w/ ST-Adapter & CLIP & 8$\times$3$\times$1 & 2062 & 86.7 & 97.5 \\
Our ViT-L w/ ST-Adapter & CLIP & 16$\times$3$\times$1 & 4124 & 86.9 & 97.6 \\
Our ViT-L w/ ST-Adapter & CLIP & 32$\times$3$\times$1 & 8248 & 87.2 & 97.6 \\
\bottomrule
\end{tabular}
\vspace{-5mm}

%% file: tables/sota_ssv2.tex
\begin{tabular}{@{}l|x{70}|x{45}|x{40}|x{45}|x{45}@{}}
\toprule
 Model & Pretrain & \#Frames & GFlops & Top-1 & Top-5\\
\toprule
\multicolumn{6}{@{}l}{\it Methods with full-finetuning} \\
TSM\citep{tsm} &IN1K & 16$\times$1$\times$1 & 66 & 63.3 & 88.5 \\
GST\citep{gst} &IN1K & 16$\times$1$\times$1 & 59 & 62.6 & 87.9 \\
MSNet\citep{msnet} & IN1K & 16$\times$1$\times$1 & 101 & 64.7 & 89.4 \\
CT-Net\citep{ct_net} & IN1K & 16$\times$1$\times$1 & 75 & 64.5 & 89.3 \\
TDN\citep{tdn} & IN1K & 16$\times$1$\times$1 & 72 & 65.3 & 89.5 \\
TimeSformer-HR\citep{timesformer} & IN21K & 16$\times$3$\times$1 & 5109 & 62.5 & - \\
X-ViT\citep{xvit} & IN21K & 32$\times$3$\times$1 & 1270 & 65.4 & 90.7 \\
Mformer-L\citep{mformer} & IN21K+K400 & 32$\times$3$\times$1 & 3555 & 68.1 & 91.2 \\
ViViT-L\citep{vivit} & IN21K+K400 & 16$\times$3$\times$4 & 11892  & 65.4 & 89.8 \\
MViT-B-24,32$\times$3\citep{fan2021multiscale} & K600 & 32$\times$1$\times$3 & 708 & 68.7 & 91.5 \\
Swin-B\citep{liu2021video} & IN21K+K400 & 32$\times$3$\times$1 & 963 & 69.6 & 92.7 \\
UniFormer-B\citep{li2022uniformer} & IN1K+K600 & 32$\times$3$\times$1 & 777 & 71.2 & 92.8 \\
\revision{OMNIVORE (Swin-B)\citep{omnivore}} & \revision{IN22K+K400+SUN} & \revision{32$\times$3$\times$1} & \revision{963} & \revision{71.4} & \revision{93.5} \\
\revision{MTV-B(320p)\citep{mtv}} & \revision{IN21K+K400} & \revision{32$\times$3$\times$4} & \revision{11160} & \revision{68.5} & \revision{90.4} \\
\revision{ViT-B w/o ST-Adapter} & \revision{CLIP} & \revision{8$\times$3$\times$1} & \revision{419} & \revision{44.0} & \revision{77.0} \\
\revision{ViT-L w/o ST-Adapter} & \revision{CLIP} & \revision{8$\times$3$\times$1} & \revision{1941} & \revision{48.7} & \revision{77.5} \\
\midrule
\multicolumn{6}{@{}l}{\it Methods with frozen backbone} \\
Our ViT-B w/ ST-Adapter & CLIP & 8$\times$3$\times$1 & 489 & 67.1 & 91.2 \\
Our ViT-B w/ ST-Adapter & CLIP & 16$\times$3$\times$1 & 977 & 69.3 & 92.3 \\
Our ViT-B w/ ST-Adapter & CLIP & 32$\times$3$\times$1 & 1955 & 69.5 & 92.6 \\
Our ViT-L w/ ST-Adapter & CLIP & 8$\times$3$\times$1 & 2062 & 70.0 & 92.3 \\
Our ViT-L w/ ST-Adapter & CLIP & 16$\times$3$\times$1 & 4124 & 71.9 & 93.4 \\
Our ViT-L w/ ST-Adapter & CLIP & 32$\times$3$\times$1 & 8248 & \textbf{72.3} & \textbf{93.9} \\
\bottomrule
\end{tabular}


%% file: tables/sota_ek100.tex
\begin{table}[ht]
    \vspace{-5mm}
    \centering
    \caption{\textbf{Results on Epic-Kitchens-100 validation set.} “Frames” denotes the total number of frames used during inference which is:  \# frames per clip $\times$ \# temporal clip $\times$ \# spatial crop.}
    \vspace{-1mm}
    \begin{tabular}{l|c|c|cc}
        \toprule
        Model & Pre-train data & \#Frames & Verb & Noun \\
        \midrule
        \multicolumn{4}{@{}l}{\it Methods with full-finetuning} \\
        ViViT-L \cite{vivit} & IN21K+K400 & 16 $\times$ 3 $\times$ 10 & 66.4 & \textbf{56.8} \\
        MFormer-B \cite{mformer} & IN21K+K400 & 16 $\times$ 3 $\times$ 10 & 66.7 & 56.5 \\
        XViT(8x) \cite{xvit} & IN21K+K400 & 8 $\times$ 3 $\times$ 1 & 66.7 & 53.3 \\
        ViT-B/16 w/o ST-Adapter & CLIP & 8 $\times$ 3 $\times$ 1 & 54.8 & 50.4 \\
        \midrule
        \multicolumn{4}{@{}l}{\it Methods with frozen backbone} \\
        Our ViT-B/16 w/ ST-Adapter & CLIP & 8 $\times$ 3 $\times$ 1 & \textbf{67.6} & 55.0 \\
        \bottomrule
    \end{tabular}
    \label{tab:epic}
\end{table}

%% file: figs/multiple_fig_in_a_row.tex
\begin{figure}[t]
    \centering
    \subfloat[Parameter Efficiency]{\label{fig:num_params}
    \includegraphics[width=0.33\textwidth]{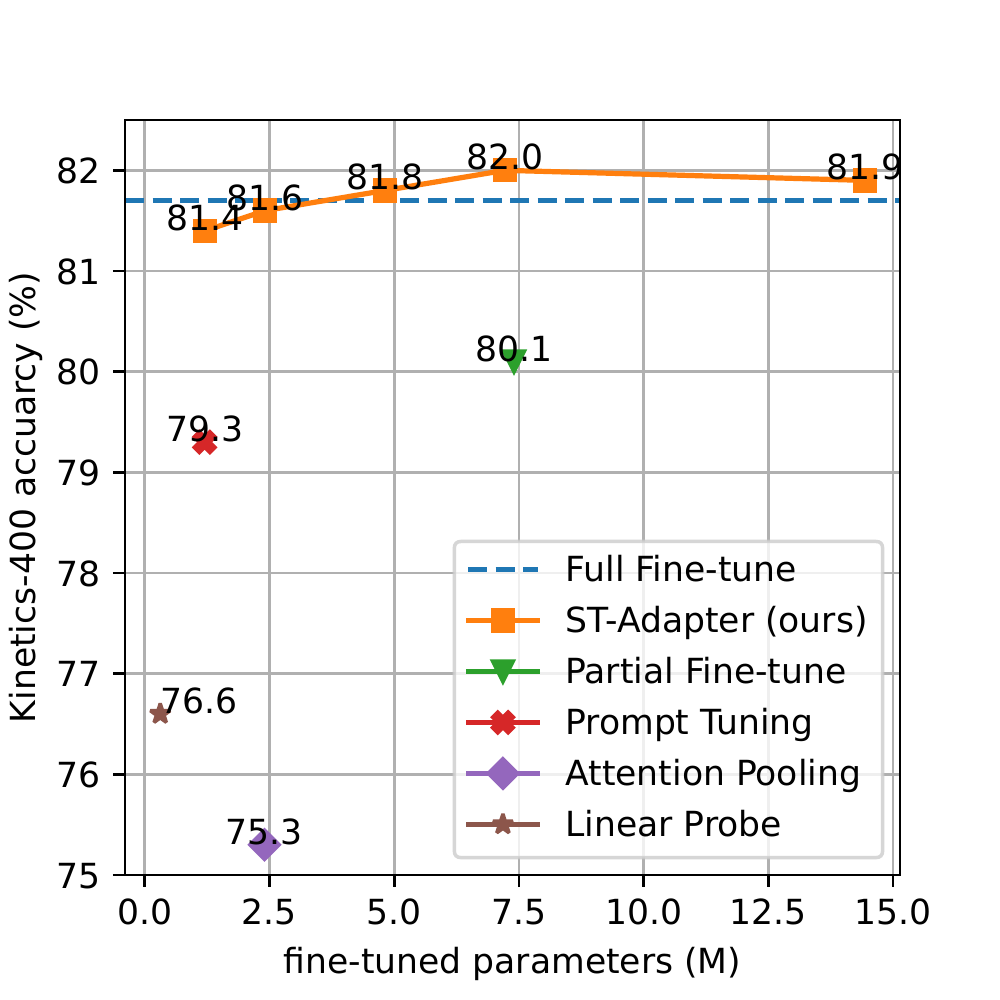}
    }
    \subfloat[Training Efficiency]{\label{fig:train_epoch}
    \includegraphics[width=0.33\textwidth]{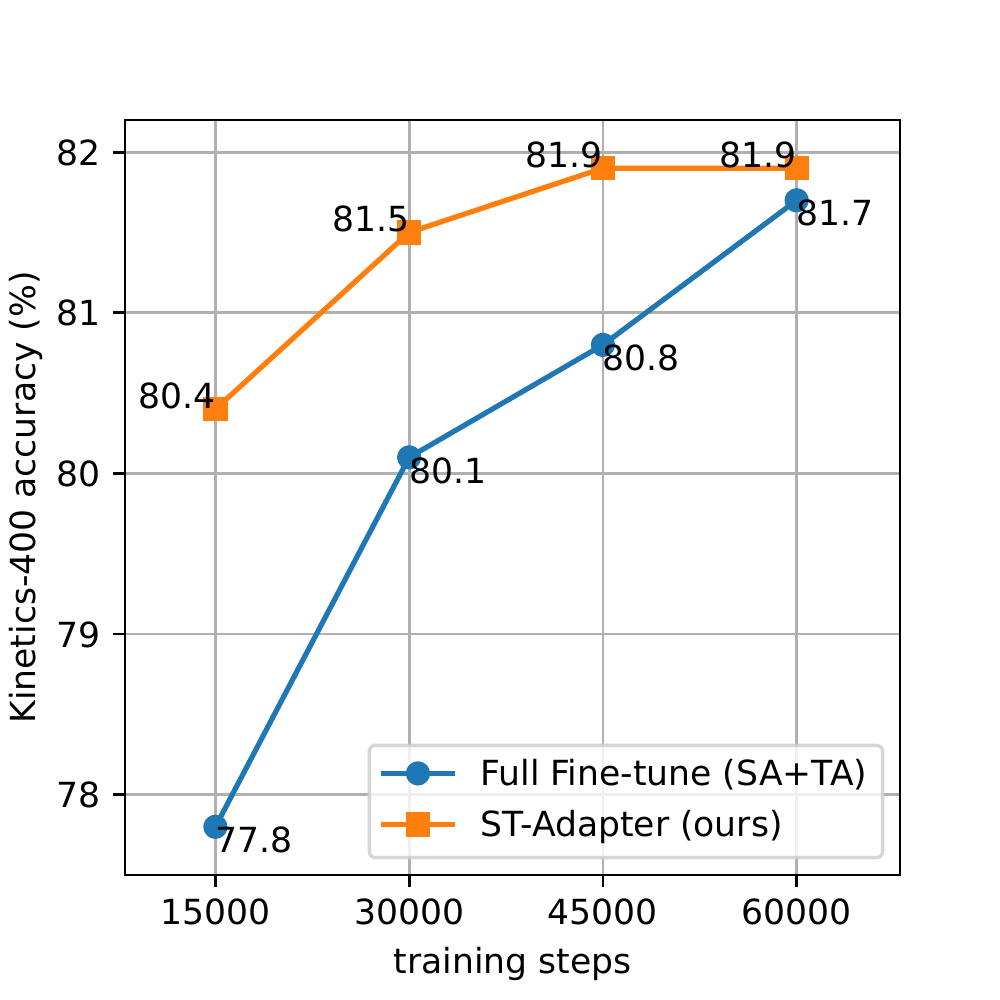}
    }
    \subfloat[Data Efficiency]{\label{fig:data_size}
    \includegraphics[width=0.33\textwidth]{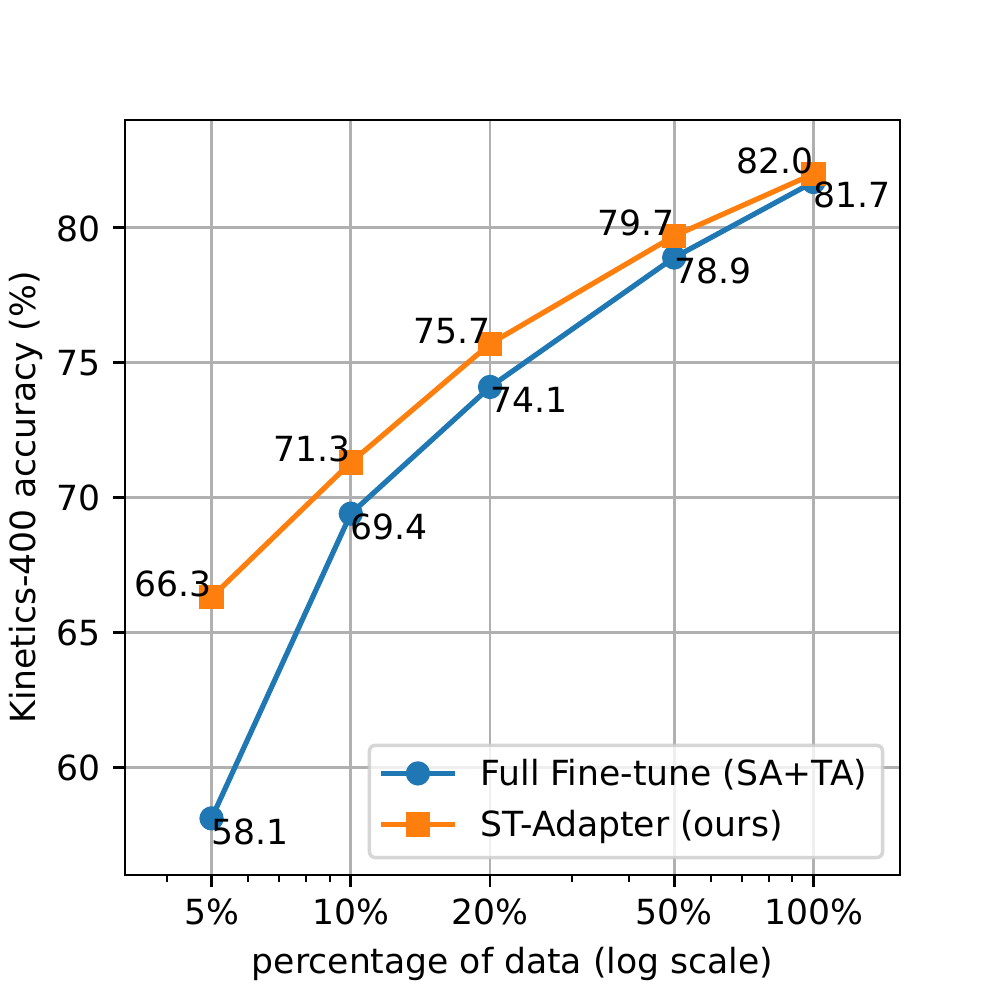}
    }
    \vspace{-2mm}
    \caption{\textbf{Ablation study on efficiency} (a) Parameter efficiency: ST-Adapter (with different bottleneck width) is compared with efficient fine-tuning methods in Table~\ref{tab:compare_efficient_transfer}.
    (b) Training efficiency: We compare ST-Adapter with Full fine-tuning under different training schedules.
    Batch size is aligned and their original schedules are shortened proportionally.
    (c) Data efficiency: Performance comparison on different training data scales. The same ViT-B/16 with CLIP pre-training is used for all experiments.}
    \vspace{-5mm}
\end{figure}

%% file: tables/ablation.tex
\begin{table*}[!t]
\vspace{-2mm}
\centering
\caption{\textbf{Ablation study on K-400 and SSv2.} (a) We show the performance with different channel numbers in the bottleneck. (b) We evenly divide the 12 blocks of ViT-B/16 into 3 groups. Block no. 1 is closest to input and no. 12 is closest to output. (c) Effect of where to put the ST-Adapter inside a block, whose diagram is shown in Fig.~\ref{fig:fig1}. 
}
\subfloat[\textbf{Bottleneck width}\label{tab:ablation:channel}]{%
\tablestyle{4.8pt}{1.1}
\tablestyle{4.8pt}{1.1}\begin{tabular}{@{}x{20}x{20}x{20}@{}}
\toprule
width & K400 & SSv2 \\
\toprule 
 64 & 81.4 & 64.4 \\ 
 128 & 81.6 & 64.9 \\
 256 & 81.8 & 65.5\\
 384 & \textbf{82.0} & \textbf{65.6} \\
 768 & 81.9 & 65.5 \\
\bottomrule
\end{tabular}}\hfill
\subfloat[\textbf{Global position} \label{tab:ablation:global_pos}]{%
\tablestyle{4.8pt}{1.1}
\tablestyle{4.8pt}{1.1}\begin{tabular}{@{}x{20}x{20}x{20}x{20}x{20}@{}}
\toprule
1-4 & 5-8 & 9-12 & K400 & SSv2 \\
\toprule 
\cmark & & & 77.7 & 45.9 \\
 & \cmark &  & 80.0 & 60.9  \\
 & & \cmark & 81.3 & 62.8 \\
 & \cmark & \cmark & 81.8 & \textbf{65.6} \\
\cmark & \cmark & \cmark & \textbf{82.0} & \textbf{65.6} \\
\bottomrule
\end{tabular}}\hfill
\subfloat[\textbf{Local position}\label{tab:ablation:local_pos}]{%
\tablestyle{4.8pt}{1.1}
\tablestyle{4.8pt}{1.1}\begin{tabular}{@{}x{60}x{20}x{20}@{}}
\toprule
position & K400 & SSv2 \\
\toprule 
before MHSA & 82.0 & 65.6 \\
after MHSA & 81.9 & 65.7 \\
after FFN & 81.9 & 65.9 \\
before \& after MHSA & 82.0 & \textbf{67.0} \\
\bottomrule
\end{tabular}}\hfill
\vspace{-5mm}
\end{table*}

%% file: tables/training_time.tex
\begin{table}[h]
    \vspace{-3mm}
    \centering
    \caption{\textbf{Training time and memory.} \revision{For full-finetuning we used the recipes in \cite{timesformer}.}}
    \begin{tabular}{l|c|c}
        \toprule
        Model & 
        \begin{tabular}{c}Training GPU-hours (K400)\end{tabular} & \begin{tabular}{c}Peak mem (MB)\end{tabular} \\
        \midrule
        TimeSformer\cite{timesformer} (Full Fine-tune)	&60 (+161\%)	& 21694 (+52\%) \\
        ViT-B/16 (Full Fine-tune)	&40 (+74\%)	&17275 (+21\%) \\
        ViT-B/16 w/ ST-Adapter	&\textbf{23}	&\textbf{14238} \\
        \bottomrule
    \end{tabular}
    \label{tab:train_time}
    \vspace{-5mm}
\end{table}

%% file: tables/kernel_shape.tex
\begin{wraptable}[10]{r}{5cm}
    \vspace{-0.5cm}
    \centering
    \caption{\textbf{Effects of kernel shape.} Kernel size is denoted as $k_T \times k_H \times k_W$ for time, height and width.}
    \begin{tabular}{ccc}
        \toprule
        Kernel Size & K400 & SSv2 \\
        \midrule
        $1 \times 1 \times 1$ & 81.6 & 46.2 \\
        $1 \times 3 \times 3$ & 81.4 & 46.2 \\
        $3 \times 1 \times 1$ & \textbf{82.0} & \textbf{66.3} \\
        $3 \times 3 \times 3$ & \textbf{82.0} & 65.6 \\
        \bottomrule
    \end{tabular}
    \label{tab:kernel_shape}
\end{wraptable}

%% file: 05-Conclusion.tex
\section{Conclusions}
In this work, we have presented a simple yet effective Spatio-Temporal Adapter (ST-Adapter)
for enabling a less studied parameter-efficient image-to-video transfer learning. 
Fully using commonly adopt primitive operators, ST-Adapter is particularly designed to be both lightweight and easy to implement for friendly usability and deployment.
This cross-modality adaptation is a practically critical capability considering that it is dramatically challenging and more costly to build a sufficiently strong large video model in reality.
Encouragingly, extensive experiments on video action recognition show that our ST-Adapter can match or surpass both 
the full fine-tuning strategy as well as fully trained state-of-the-art video models, 
whilst having the benefit of (20 times less updated parameters) parameter-efficiency.
Further, our method is also faster to train
and consumes less computing resources with economic and environmental superiority.
We believe this work is inspiring for the research of other video understanding tasks such as action localization and video summarization. 

{\flushleft \bf Acknowledgement} This work is supported in part by Centre for Perceptual and Interactive Intelligence Limited, in part by the General Research Fund through the Research Grants Council of Hong Kong under Grants (Nos. 14204021, 14207319).

%% file: checklist.tex
\section*{Checklist}

\begin{enumerate}

\item For all authors...
\begin{enumerate}
  \item Do the main claims made in the abstract and introduction accurately reflect the paper's contributions and scope?
    \answerYes{}
  \item Did you describe the limitations of your work?
    \answerNo{}
  \item Did you discuss any potential negative societal impacts of your work?
    \answerNo{}
  \item Have you read the ethics review guidelines and ensured that your paper conforms to them?
    \answerYes{}
\end{enumerate}

\item If you are including theoretical results...
\begin{enumerate}
  \item Did you state the full set of assumptions of all theoretical results?
    \answerNA{}
  \item Did you include complete proofs` of all theoretical results?
    \answerNA{}
\end{enumerate}

\item If you ran experiments...
\begin{enumerate}
  \item Did you include the code, data, and instructions needed to reproduce the main experimental results (either in the supplemental material or as a URL)?
    \answerYes{Code will be provided on GitHub after blind review.}
  \item Did you specify all the training details (e.g., data splits, hyperparameters, how they were chosen)?
    \answerYes{See \ref{sec:Appendix}}
  \item Did you report error bars (e.g., with respect to the random seed after running experiments multiple times)?
    \answerNo{The experiments are too expensive to repeat many times.}
  \item Did you include the total amount of compute and the type of resources used (e.g., type of GPUs, internal cluster, or cloud provider)?
    \answerYes{See \ref{sec:Appendix}}
\end{enumerate}

\item If you are using existing assets (e.g., code, data, models) or curating/releasing new assets...
\begin{enumerate}
  \item If your work uses existing assets, did you cite the creators?
    \answerYes{All are mentioned in \ref{sec:Experiments}}
  \item Did you mention the license of the assets?
    \answerNo{}
  \item Did you include any new assets either in the supplemental material or as a URL?
    \answerYes{Code will be provided on GitHub after blind review.}
  \item Did you discuss whether and how consent was obtained from people whose data you're using/curating?
    \answerNo{}
  \item Did you discuss whether the data you are using/curating contains personally identifiable information or offensive content?
    \answerNo{}
\end{enumerate}

\item If you used crowdsourcing or conducted research with human subjects...
\begin{enumerate}
  \item Did you include the full text of instructions given to participants and screenshots, if applicable?
    \answerNA{}
  \item Did you describe any potential participant risks, with links to Institutional Review Board (IRB) approvals, if applicable?
    \answerNA{}
  \item Did you include the estimated hourly wage paid to participants and the total amount spent on participant compensation?
    \answerNA{}
\end{enumerate}

\end{enumerate}

%% file: 06-Appendix.tex
\appendix
\section{Appendix}
\label{sec:Appendix}

{\flushleft \bf Implementation details of our method}~\label{impl_details} All experiments are implemented in PyTorch \cite{paszke2019pytorch}. We use the configuration listed in Tab.~\ref{tab:impl_detail} unless otherwise specified. In general, we use much simpler data augmentation techniques compared to end-to-end fine-tuning.
\revision{
Hyper-parameters were briefly tuned to ensure convergence on a 20\% held-out validation set.
}

\begin{table}[h]
    \centering
    \caption{{\bf Default implementation details of our method.}}
    \begin{tabular}{lcccc}
        \toprule
        dataset and backbone & K400, ViT-B & K400, ViT-L & SSv2, ViT-B & SSv2, ViT-L \\
        \midrule
        num. adapters per block & 1 & 1 & 2 & 1 \\
        adapter bottleneck width & \multicolumn{4}{c}{384} \\
        convolution kernel shape & \multicolumn{4}{c}{$3 \times 1 \times 1$ ($k_T \times k_H \times k_W$)} \\
        \midrule
        optimizer & \multicolumn{4}{c}{AdamW, learning rate=5e-4, weight decay=1e-2} \\
        batch size & \multicolumn{4}{c}{128} \\
        training steps & 20k & 40k & 50k & 50k \\
        training resize & \begin{tabular}{c}ShortSideJitter\\224 - 256\end{tabular} & \multicolumn{3}{c}{RandomResizedCrop} \\
        training crop size & \multicolumn{4}{c}{224} \\
        frame sampling rate &
        \multicolumn{2}{c}{
        \begin{tabular}{c}16 (for 8 frames per view)\\8 (for 16 frames per view)\\4 (for 32 frames per view)\end{tabular}
        } &
        \multicolumn{2}{c}{\begin{tabular}{c}dynamic, evenly covering\\the whole video\end{tabular}
        } \\
        mirror & \cmark & \cmark & \xmark & \xmark \\
        RandAugment \cite{cubuk2020randaugment} & \xmark & \xmark & \cmark & \cmark \\
        \midrule
        num. testing views & \multicolumn{2}{c}{3 temporal $\times$ 1 spatial} & \multicolumn{2}{c}{1 temporal $\times$ 3 spatial} \\
        \bottomrule
    \end{tabular}
    \label{tab:impl_detail}
\end{table}

{\flushleft \bf Baseline implementation details}~The training configuration used for all the baselines is summarized as follows:

\begin{itemize}
    \item {\it Full Fine-tuning}: we largely follow the training configuration provided in their original paper, except that we train all the CLIP initialized layers with 1/100 learning rate and weight decay. We found these changes are necessary to obtain reasonable results for CLIP pretrained models; Otherwise the accuracy on Kinetics-400 is less than 50\%.
    We found 1/100 to be the best scaling among $\{1/10, 1/100, 1/1000\}$ on Kinetics-400.
    
    \item {\it Partial Fine-tuning}: we finetune only the last Transformer block and the classifier layer. For the SA+TA architecture, TA is only added to the last block since the previous blocks need to be frozen in a meaningful state. We use the identical training configuration as provided in the original paper ({\it i.e.}, without reduction of learning rate or weight decay for any trainable weight) as we found it slightly improves accuracy for this baseline.
    
    \item Other baselines use the same training configuration as our proposed method, as stated in the {\it Implementation details} section in the main manuscript.
\end{itemize}

{\flushleft \revision{\bf Experiments with other foundation models}} \revision{It is observed that with the same model, CLIP pre-training is superior to ImageNet21K pre-training (not surprising due to the training data scale and richness difference). However, our main objective is to propose a parameter-efficient fine-tuning alternative to the standard full fine-tuning approach particularly for image-to-video adaptation. To that end, we have validated the effectiveness and efficiency of turning an image foundation model into strong video action recognition models by tuning only a small fraction of parameters, in comparison to previous state-of-the-art alternatives.}

\revision{By reporting the results on two different pre-training datasets (i.e., ImageNet21K and CLIP datasets), we would like to demonstrate that our ST-Adapter can generalize across different pre-training datasets and methods. Moreover, it can shed light on the difference between a foundation model (pre-trained with noisy web-scale raw data) and an ImageNet pre-trained model (which has been standard pre-training over the last decade).}

\revision{To further support our finding, we have also experimented with the latest SWAG \cite{swag} foundation model. As seen in Table \ref{tab:swag}, our ST-Adapter with a SWAG model can achieve consistent results as with a CLIP model: Reaching similar accuracy in the same tendency whilst outperforming the strong full fine-tuning strategy on both action datasets.}
\input{appendix_files/swag}

{\flushleft \revision{\bf Experiments on additional backbone architectures}} \revision{
we have additionally provided the results of ST-Adapters on Swin-B models in Table~\ref{tab:swin}. The results of Swin space only and Swin joint attention are obtained with the training configure of \cite{liu2021video} but using (8 frames x 3 views) sampling setting. Although they are not directly comparable with the results reported in \cite{liu2021video} (32 frames $\times$ 12 views for K400, 32 frames $\times$ 3 views for SSv2), they are highly indicative within reasonable range. It is expected that on ImageNet-21k pretrained models our ST-Adapters underperforms full fine-tuning, especially when the locality inductive bias of Swin makes tuning on the downstream tasks easier. However, our ST-Adapter still exhibits strong temporal learning capability, matching the joint-attention Swin and outperforms space-only Swin by a large margin. Also, we observe higher data efficiency with our ST-Adapter: The Swin joint attention model on the SSv2 dataset relies on K400 pretraining (directly fine-tuning from ImageNet-21k results in slightly less than 60\% accuracy). In contrast, Swin w/ ST-Adapter achieves 65.1\% even when directly trained from ImageNet-21k weights. 
Note, we primarily aim at adapting foundation image models\cite{foundation} pretrained on larger datasets (e.g., CLIP) other than ImageNet-21k.
}

\input{appendix_files/swin}

{\flushleft \revision{\bf Inference Speed}} \revision{We provide an inference speed test in Table~\ref{tab:speed}. We measure the latency at batch size = 1 and throughput at batch size = 32. It is shown that our model performs slightly lower than TimeSformer space only, indicating that just a small overhead is introduced in inference speed by ST-Adapter.
}
\input{appendix_files/speed}

{\flushleft \bf UCF-101 and HMDB-51}~We verify our method on two additional smaller but also widely studied video recognition datasets, namely UCF-101 \cite{ucf101} and HMDB-51 \cite{hmdb}. For both cases, we finetune from a Kinetics-400~\cite{i3d} pretrained model, with all CLIP layers fixed and ST-Adapters set to 1/10 learning rate and weight decay, and train for 500 steps with a batch size of 128. Frames are sampled with a temporal stride of 8. All other training settings are identical to that used for Kinetics-400. For testing, we use 3 spatial views and 2 temporal views, and report the 3-split mean accuracy for both datasets. We compare with methods that take only RGB frames as input (without optical flow).
The results are shown in Table~\ref{tab:ucf_hmdb}. 
We observe similar top performance by our ST-Adapter
in comparison to recent state-of-the-art competitors,
including the latest CLIP based VideoPrompt by a large margin.

\input{appendix_files/res_ucf101} 

\begin{figure}[t]
    \centering
    \subfloat{\includegraphics[width=0.33\textwidth]{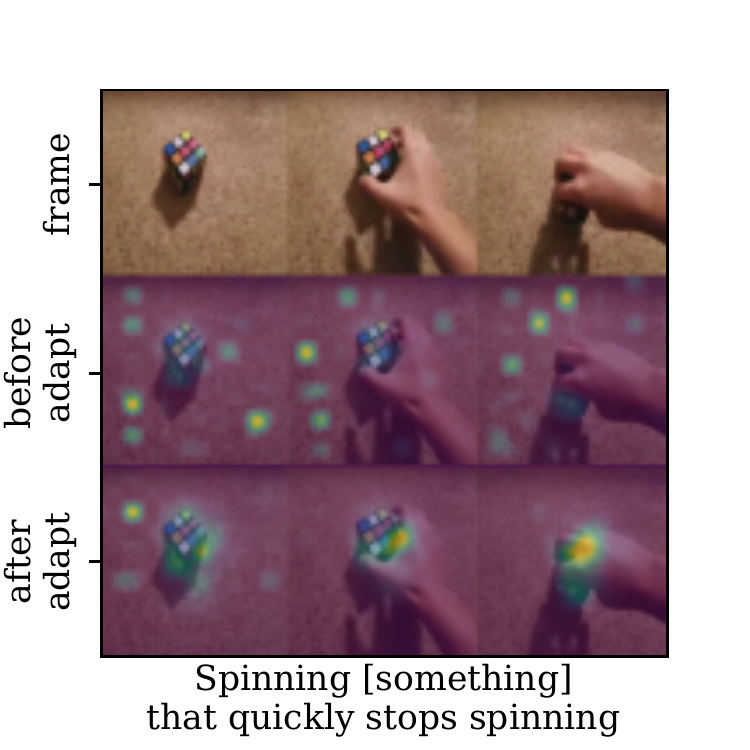}}
    \subfloat{\includegraphics[width=0.33\textwidth]{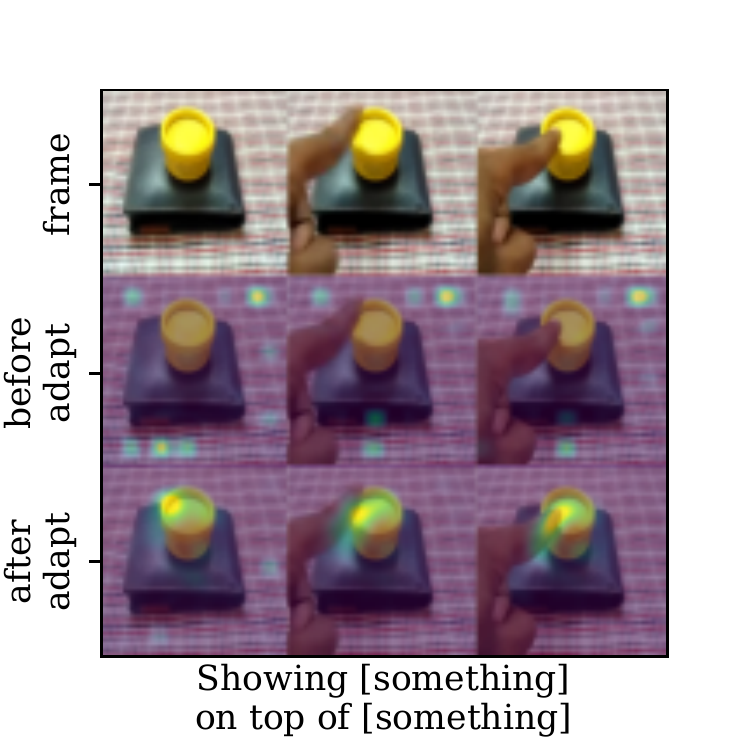}}
    \subfloat{\includegraphics[width=0.33\textwidth]{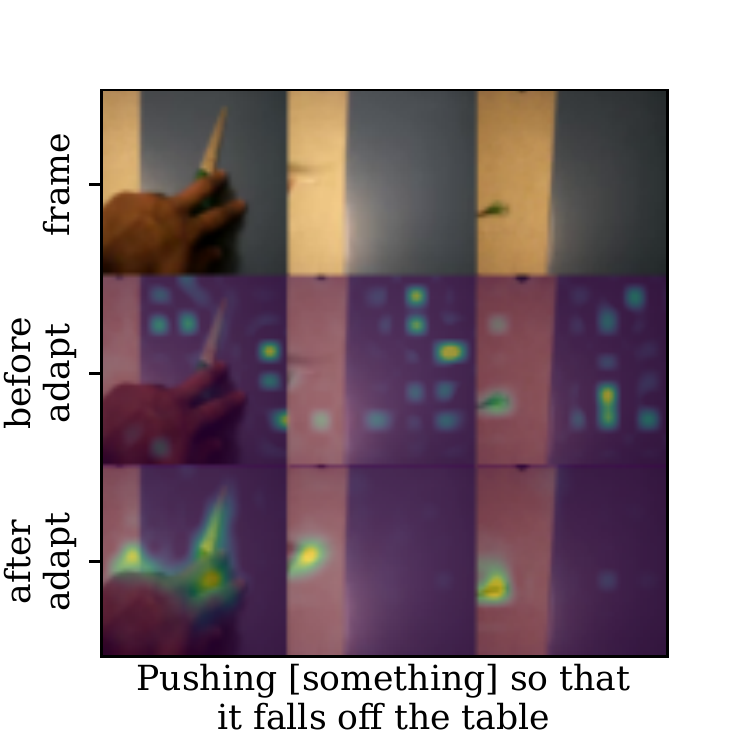}}
    \\
    \subfloat{\includegraphics[width=0.33\textwidth]{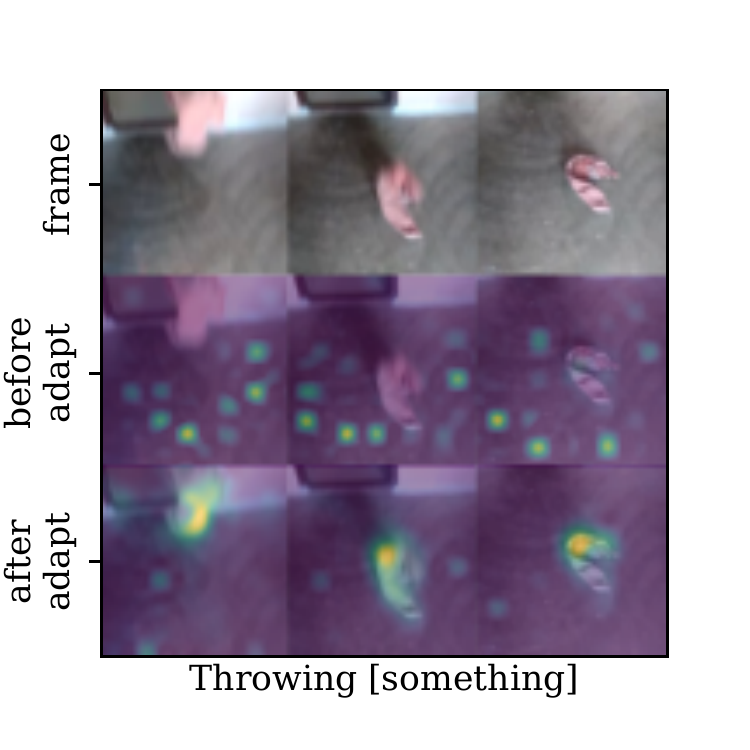}}
    \subfloat{\includegraphics[width=0.33\textwidth]{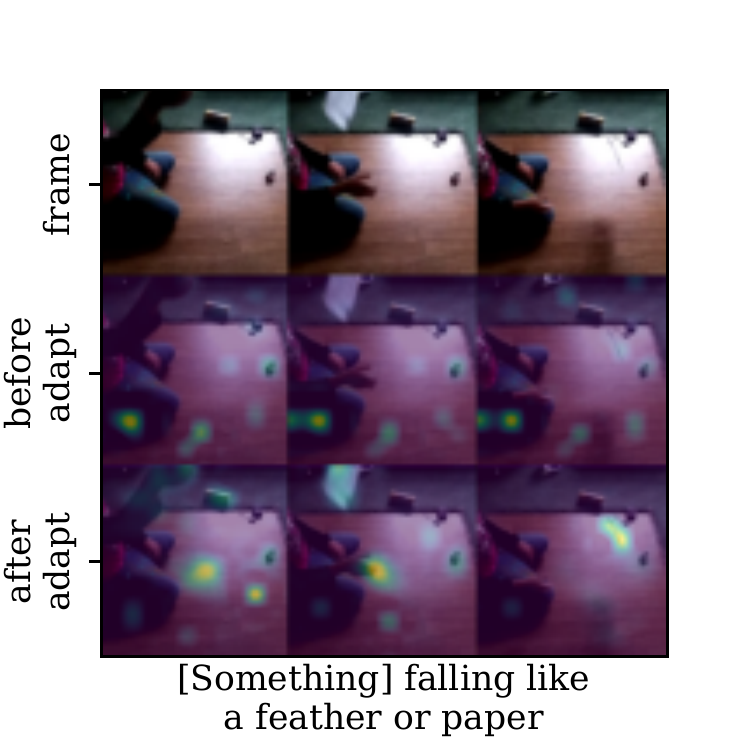}}
    \subfloat{\includegraphics[width=0.33\textwidth]{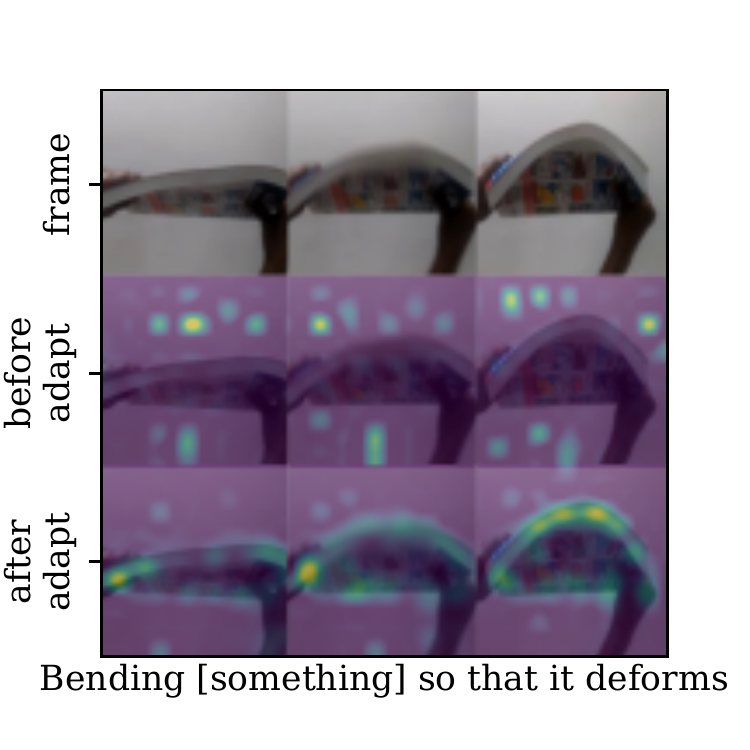}}
    \\
    \subfloat{\includegraphics[width=0.33\textwidth]{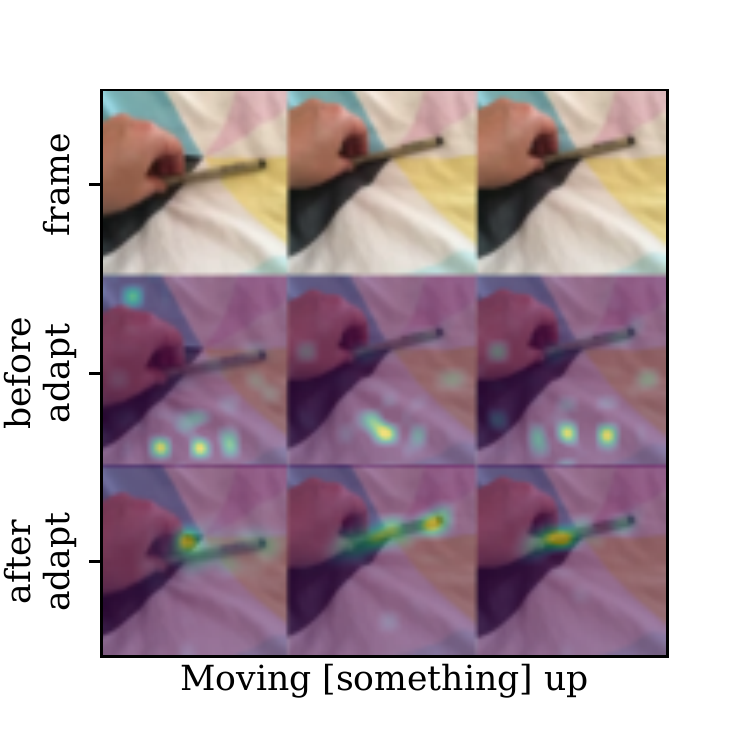}}
    \subfloat{\includegraphics[width=0.33\textwidth]{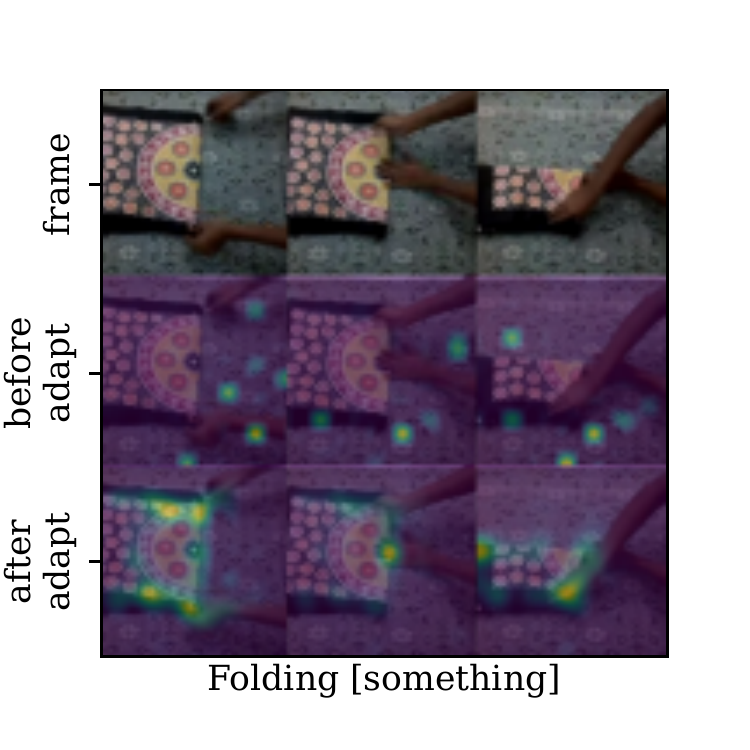}}
    \subfloat{\includegraphics[width=0.33\textwidth]{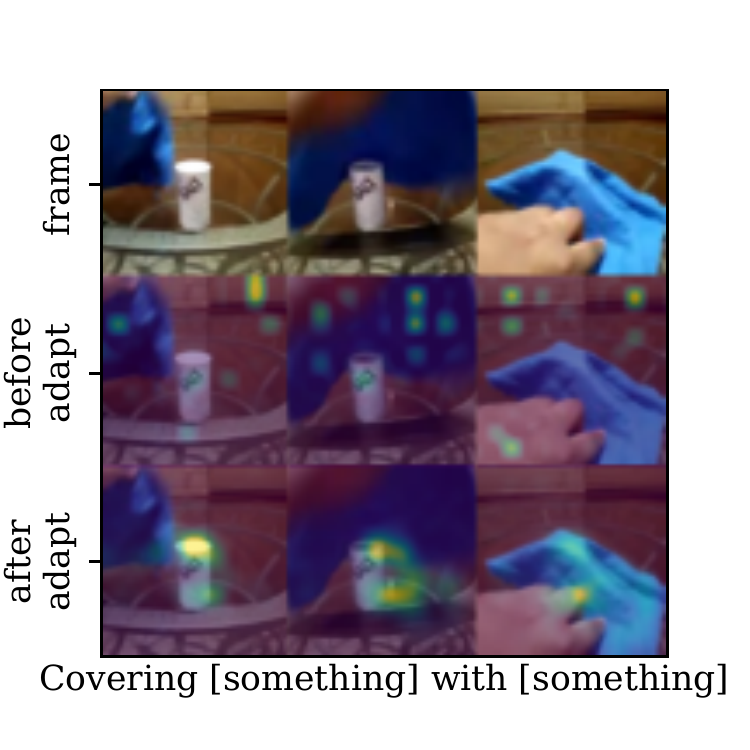}}
    \\
    \subfloat{\includegraphics[width=0.33\textwidth]{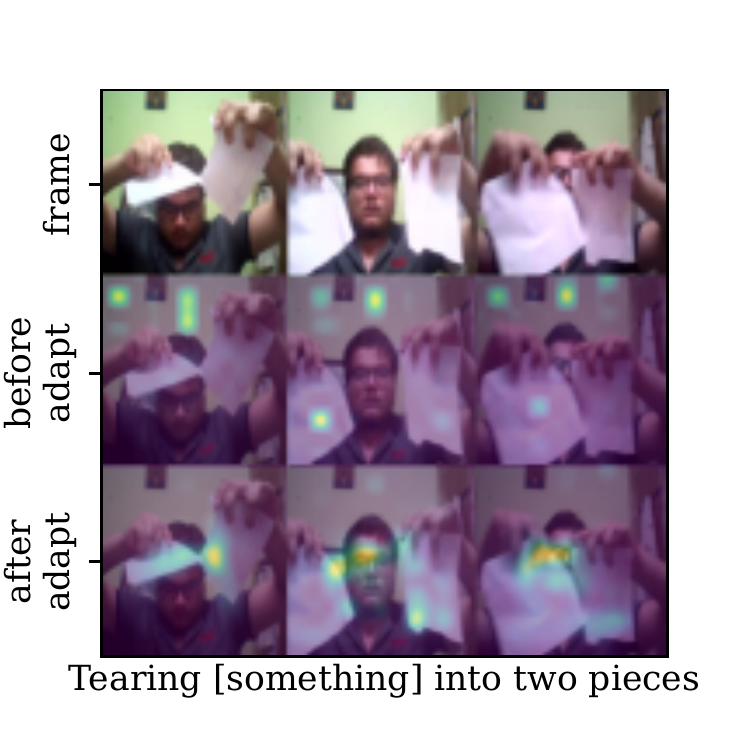}}
    \subfloat{\includegraphics[width=0.33\textwidth]{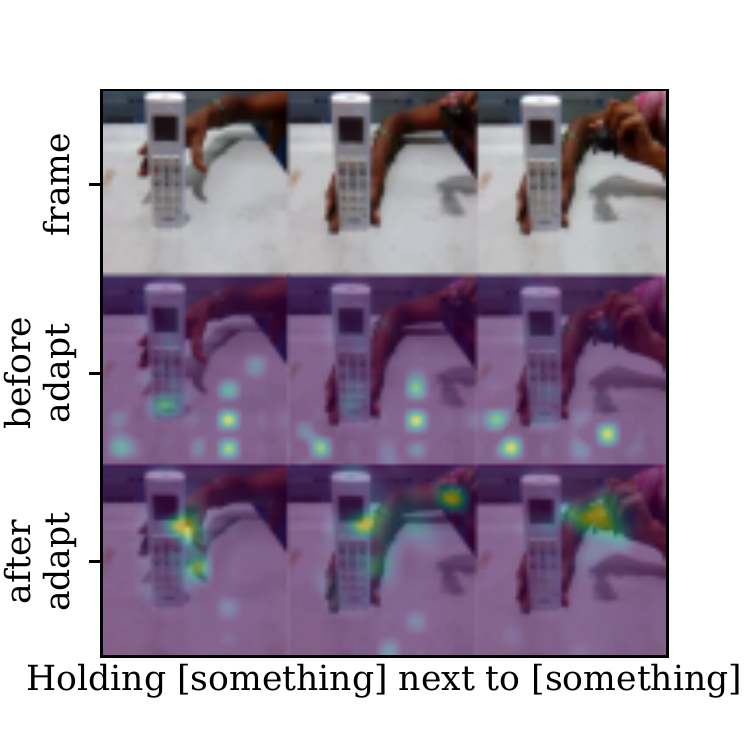}}
    \subfloat{\includegraphics[width=0.33\textwidth]{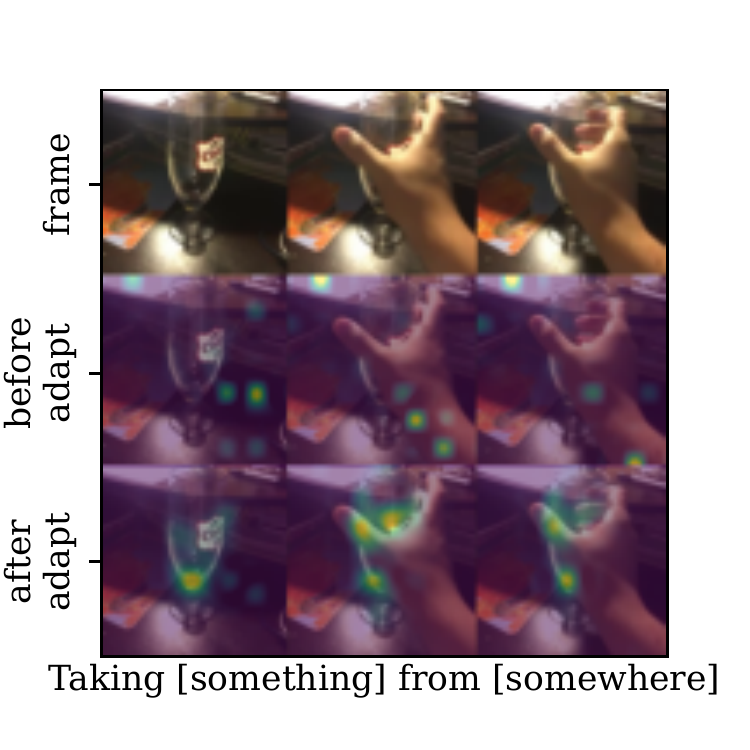}}
    \\
    \caption{{\bf Visualization of attention map before and after ST-Adaptation.}}
    \label{fig:attnmap}
\end{figure}

{\flushleft \bf Visualization}~We provide qualitative results about the attention map change before and after adding the ST-Adapters in Fig.~\ref{fig:attnmap}. Videos are sampled from Something-Something-v2 dataset and the attention map of the \verb|[CLS]| token from the last Transformer block is shown. The visualization shows that with ST-Adapters, the model attends more to 
action related regions ({\em e.g.}, hands, fore-ground objects or moving objects), while the CLIP model without adaptation tend to be distracted by the background.

%% file: appendix_files/swag.tex
\begin{table}[h]
    \centering
    \caption{\textbf{Experiment with different foundation models.}}
    \begin{tabular}{l|c|c|c}
        \toprule
        Model & Pre-train & K400 & SSv2 \\
        \midrule
        ViT-B (Full Fine-tuning) & CLIP & 81.0 & 44.0\\
        ViT-B w/ ST-Adapter	& CLIP&	82.0&	67.1\\
        ViT-B (Full Fine-tuning)	&SWAG&	80.1&	45.2\\
        ViT-B w/ ST-Adapter	&SWAG&	80.9&	67.2\\
        \bottomrule
    \end{tabular}
    \label{tab:swag}
\end{table}

%% file: appendix_files/swin.tex
\begin{table}[h]
    \centering
    \caption{\textbf{Experiment with SWIN Transformers.}}
    \begin{tabular}{l|c|c}
        \toprule
        Model & K400 & SSv2 \\
        \midrule
        Swin SpaceOnly (Full Fine-tuning) & 80.1 & 44.3 \\
        Swin Join-Attention (Full Fine-tuning) & 81.5 & 65.3 \\
        \midrule
        Swin w/ ST-Adapter & 77.1 & 65.1 \\
        \bottomrule
    \end{tabular}
    \label{tab:swin}
\end{table}

%% file: appendix_files/speed.tex
\begin{table}[h]
    \centering
    \caption{\textbf{Inference Speed.}}
    \begin{tabular}{l|c|c|c|c}
        \toprule
        Model & \begin{tabular}{c}Total number of \\Params (M)\end{tabular}  & K400 & \begin{tabular}{c}Latency\\(ms)\end{tabular} & \begin{tabular}{c}Throughput\\(V/s)\end{tabular}  \\
        \midrule
        TimeSformer\cite{timesformer}	&121.57	&81.7	&28	&69 \\
        ViT-B/16 &	86.11	&81.0	&17	&98 \\
        ViT-B/16 w/ ST-Adapter	&93.00	&82.0	&19	&90 \\
        \bottomrule
    \end{tabular}
    \label{tab:speed}
\end{table}

%% file: appendix_files/res_ucf101.tex
\begin{table}[h]
    \centering
    \caption{\textbf{Comparing the state-of-the-art video recognition methods on UCF101 and HMDB51.}}
    \begin{tabular}{l|c|c|c}
        \toprule
        Method & Pre-train data & UCF101 & HMDB51 \\
        \midrule
        STC \cite{stc} & K400 & 95.8 & 72.6 \\
        ECO \cite{eco} & K400 & 93.6 & 68.4 \\
        R(2+1)D-34 \cite{r21} & K400 & 96.8 & 74.5 \\
        I3D \cite{i3d} & ImageNet+K400 & 95.6 & 74.8 \\
        S3D \cite{s3d} & ImageNet+K400 & 96.8 & 75.9 \\
        FASTER32 \cite{faster} & K400 & 96.9 & 75.7 \\
        VideoPrompt \cite{vidprompt} & CLIP & 93.6 & 66.4 \\
        SlowOnly-8x8-R101 \cite{omni} & Kinetics+OmniSource\cite{omni} & 97.3 & 79.0 \\
        \midrule
        ViT-B/16 w/ ST-Adapter (Ours) & CLIP+K400 & 96.4 & 77.7 \\
        ViT-L/14 w/ ST-Adapter (Ours) & CLIP+K400 & 98.1 & 81.7 \\
        ViT-L/14@336px w/ ST-Adapter (Ours) & CLIP+K400 & \textbf{98.3} & \textbf{82.8} \\
        \bottomrule
    \end{tabular}
    \label{tab:ucf_hmdb}
\end{table}

%% file: main.bbl
\begin{thebibliography}{96}
\providecommand{\natexlab}[1]{#1}
\providecommand{\url}[1]{\texttt{#1}}
\expandafter\ifx\csname urlstyle\endcsname\relax
  \providecommand{\doi}[1]{doi: #1}\else
  \providecommand{\doi}{doi: \begingroup \urlstyle{rm}\Url}\fi

\bibitem[Akbari et~al.(2021)Akbari, Yuan, Qian, Chuang, Chang, Cui, and
  Gong]{vatt}
Hassan Akbari, Liangzhe Yuan, Rui Qian, Wei-Hong Chuang, Shih-Fu Chang, Yin
  Cui, and Boqing Gong.
\newblock Vatt: Transformers for multimodal self-supervised learning from raw
  video, audio and text.
\newblock \emph{Advances in Neural Information Processing Systems}, 34, 2021.

\bibitem[Arnab et~al.(2021)Arnab, Dehghani, Heigold, Sun, Lu{\v{c}}i{\'c}, and
  Schmid]{vivit}
Anurag Arnab, Mostafa Dehghani, Georg Heigold, Chen Sun, Mario Lu{\v{c}}i{\'c},
  and Cordelia Schmid.
\newblock Vivit: A video vision transformer.
\newblock In \emph{Proceedings of the IEEE/CVF International Conference on
  Computer Vision}, pages 6836--6846, 2021.

\bibitem[Ba et~al.(2016)Ba, Kiros, and Hinton]{ba2016layer}
Jimmy~Lei Ba, Jamie~Ryan Kiros, and Geoffrey~E Hinton.
\newblock Layer normalization.
\newblock \emph{arXiv preprint arXiv:1607.06450}, 2016.

\bibitem[Bahng et~al.(2022)Bahng, Jahanian, Sankaranarayanan, and Isola]{vp}
Hyojin Bahng, Ali Jahanian, Swami Sankaranarayanan, and Phillip Isola.
\newblock Visual prompting: Modifying pixel space to adapt pre-trained models.
\newblock \emph{arXiv preprint arXiv:2203.17274}, 2022.

\bibitem[Bapna et~al.(2019)Bapna, Arivazhagan, and Firat]{bapna2019simple}
Ankur Bapna, Naveen Arivazhagan, and Orhan Firat.
\newblock Simple, scalable adaptation for neural machine translation.
\newblock \emph{arXiv preprint arXiv:1909.08478}, 2019.

\bibitem[Bertasius et~al.(2021)Bertasius, Wang, and Torresani]{timesformer}
Gedas Bertasius, Heng Wang, and Lorenzo Torresani.
\newblock Is space-time attention all you need for video understanding?
\newblock \emph{arXiv preprint arXiv:2102.05095}, 2021.

\bibitem[Bommasani et~al.(2021)Bommasani, Hudson, Adeli, Altman, Arora, von
  Arx, Bernstein, Bohg, Bosselut, Brunskill, et~al.]{foundation}
Rishi Bommasani, Drew~A Hudson, Ehsan Adeli, Russ Altman, Simran Arora, Sydney
  von Arx, Michael~S Bernstein, Jeannette Bohg, Antoine Bosselut, Emma
  Brunskill, et~al.
\newblock On the opportunities and risks of foundation models.
\newblock \emph{arXiv preprint arXiv:2108.07258}, 2021.

\bibitem[Brown et~al.(2020)Brown, Mann, Ryder, Subbiah, Kaplan, Dhariwal,
  Neelakantan, Shyam, Sastry, Askell, Agarwal, Herbert-Voss, Krueger, Henighan,
  Child, Ramesh, Ziegler, Wu, Winter, Hesse, Chen, Sigler, Litwin, Gray, Chess,
  Clark, Berner, McCandlish, Radford, Sutskever, and Amodei]{gpt3}
Tom Brown, Benjamin Mann, Nick Ryder, Melanie Subbiah, Jared~D Kaplan, Prafulla
  Dhariwal, Arvind Neelakantan, Pranav Shyam, Girish Sastry, Amanda Askell,
  Sandhini Agarwal, Ariel Herbert-Voss, Gretchen Krueger, Tom Henighan, Rewon
  Child, Aditya Ramesh, Daniel Ziegler, Jeffrey Wu, Clemens Winter, Chris
  Hesse, Mark Chen, Eric Sigler, Mateusz Litwin, Scott Gray, Benjamin Chess,
  Jack Clark, Christopher Berner, Sam McCandlish, Alec Radford, Ilya Sutskever,
  and Dario Amodei.
\newblock Language models are few-shot learners.
\newblock volume~33, pages 1877--1901. Curran Associates, Inc., 2020.

\bibitem[Bulat et~al.(2021)Bulat, Perez~Rua, Sudhakaran, Martinez, and
  Tzimiropoulos]{xvit}
Adrian Bulat, Juan~Manuel Perez~Rua, Swathikiran Sudhakaran, Brais Martinez,
  and Georgios Tzimiropoulos.
\newblock Space-time mixing attention for video transformer.
\newblock \emph{Advances in Neural Information Processing Systems}, 34, 2021.

\bibitem[Carreira and Zisserman(2017)]{i3d}
Joao Carreira and Andrew Zisserman.
\newblock Quo vadis, action recognition? a new model and the kinetics dataset.
\newblock In \emph{proceedings of the IEEE Conference on Computer Vision and
  Pattern Recognition}, pages 6299--6308, 2017.

\bibitem[Carreira et~al.(2018)Carreira, Noland, Banki-Horvath, Hillier, and
  Zisserman]{carreira2018short}
Joao Carreira, Eric Noland, Andras Banki-Horvath, Chloe Hillier, and Andrew
  Zisserman.
\newblock A short note about kinetics-600.
\newblock \emph{arXiv preprint arXiv:1808.01340}, 2018.

\bibitem[Carreira et~al.(2019)Carreira, Noland, Hillier, and
  Zisserman]{carreira2019short}
Joao Carreira, Eric Noland, Chloe Hillier, and Andrew Zisserman.
\newblock A short note on the kinetics-700 human action dataset.
\newblock \emph{arXiv preprint arXiv:1907.06987}, 2019.

\bibitem[Cubuk et~al.(2020)Cubuk, Zoph, Shlens, and Le]{cubuk2020randaugment}
Ekin~D Cubuk, Barret Zoph, Jonathon Shlens, and Quoc~V Le.
\newblock Randaugment: Practical automated data augmentation with a reduced
  search space.
\newblock In \emph{Proceedings of the IEEE/CVF Conference on Computer Vision
  and Pattern Recognition Workshops}, pages 702--703, 2020.

\bibitem[Damen et~al.(2022)Damen, Doughty, Farinella, , Furnari, Ma, Kazakos,
  Moltisanti, Munro, Perrett, Price, and Wray]{epic}
Dima Damen, Hazel Doughty, Giovanni~Maria Farinella, , Antonino Furnari, Jian
  Ma, Evangelos Kazakos, Davide Moltisanti, Jonathan Munro, Toby Perrett, Will
  Price, and Michael Wray.
\newblock Rescaling egocentric vision: Collection, pipeline and challenges for
  epic-kitchens-100.
\newblock \emph{International Journal of Computer Vision (IJCV)}, 130:\penalty0
  33–55, 2022.
\newblock URL \url{https://doi.org/10.1007/s11263-021-01531-2}.

\bibitem[Deng et~al.(2009)Deng, Dong, Socher, Li, Li, and
  Fei-Fei]{imagenet_cvpr09}
Jia Deng, Wei Dong, Richard Socher, Li-Jia Li, Kai Li, and Li~Fei-Fei.
\newblock Imagenet: A large-scale hierarchical image database.
\newblock 2009.

\bibitem[Devlin et~al.(2019)Devlin, Chang, Lee, and Toutanova]{bert}
Jacob Devlin, Ming-Wei Chang, Kenton Lee, and Kristina Toutanova.
\newblock {BERT}: Pre-training of deep bidirectional transformers for language
  understanding.
\newblock In \emph{Proceedings of the 2019 Conference of the North {A}merican
  Chapter of the Association for Computational Linguistics: Human Language
  Technologies, Volume 1 (Long and Short Papers)}, pages 4171--4186,
  Minneapolis, Minnesota, June 2019. Association for Computational Linguistics.

\bibitem[Diba et~al.(2018)Diba, Fayyaz, Sharma, Arzani, Yousefzadeh, Gall, and
  Van~Gool]{stc}
Ali Diba, Mohsen Fayyaz, Vivek Sharma, M~Mahdi Arzani, Rahman Yousefzadeh,
  Juergen Gall, and Luc Van~Gool.
\newblock Spatio-temporal channel correlation networks for action
  classification.
\newblock In \emph{Proceedings of the European Conference on Computer Vision
  (ECCV)}, pages 284--299, 2018.

\bibitem[Dosovitskiy et~al.(2020)Dosovitskiy, Beyer, Kolesnikov, Weissenborn,
  Zhai, Unterthiner, Dehghani, Minderer, Heigold, Gelly, et~al.]{vit}
Alexey Dosovitskiy, Lucas Beyer, Alexander Kolesnikov, Dirk Weissenborn,
  Xiaohua Zhai, Thomas Unterthiner, Mostafa Dehghani, Matthias Minderer, Georg
  Heigold, Sylvain Gelly, et~al.
\newblock An image is worth 16x16 words: Transformers for image recognition at
  scale.
\newblock 2020.

\bibitem[Duan et~al.(2020)Duan, Zhao, Xiong, Liu, and Lin]{omni}
Haodong Duan, Yue Zhao, Yuanjun Xiong, Wentao Liu, and Dahua Lin.
\newblock Omni-sourced webly-supervised learning for video recognition.
\newblock In \emph{European Conference on Computer Vision}, pages 670--688.
  Springer, 2020.

\bibitem[Fan et~al.(2021)Fan, Xiong, Mangalam, Li, Yan, Malik, and
  Feichtenhofer]{fan2021multiscale}
Haoqi Fan, Bo~Xiong, Karttikeya Mangalam, Yanghao Li, Zhicheng Yan, Jitendra
  Malik, and Christoph Feichtenhofer.
\newblock Multiscale vision transformers.
\newblock \emph{arXiv preprint arXiv:2104.11227}, 2021.

\bibitem[Feichtenhofer(2020)]{feichtenhofer2020x3d}
Christoph Feichtenhofer.
\newblock X3d: Expanding architectures for efficient video recognition.
\newblock In \emph{Proceedings of the IEEE/CVF Conference on Computer Vision
  and Pattern Recognition}, pages 203--213, 2020.

\bibitem[Feichtenhofer et~al.(2019)Feichtenhofer, Fan, Malik, and
  He]{feichtenhofer2019slowfast}
Christoph Feichtenhofer, Haoqi Fan, Jitendra Malik, and Kaiming He.
\newblock Slowfast networks for video recognition.
\newblock In \emph{Proceedings of the IEEE/CVF International Conference on
  Computer Vision}, pages 6202--6211, 2019.

\bibitem[Feichtenhofer et~al.(2021)Feichtenhofer, Fan, Xiong, Girshick, and
  He]{feichtenhofer2021large}
Christoph Feichtenhofer, Haoqi Fan, Bo~Xiong, Ross Girshick, and Kaiming He.
\newblock A large-scale study on unsupervised spatiotemporal representation
  learning.
\newblock In \emph{Proceedings of the IEEE/CVF Conference on Computer Vision
  and Pattern Recognition}, pages 3299--3309, 2021.

\bibitem[Girdhar et~al.(2022)Girdhar, Singh, Ravi, van~der Maaten, Joulin, and
  Misra]{omnivore}
Rohit Girdhar, Mannat Singh, Nikhila Ravi, Laurens van~der Maaten, Armand
  Joulin, and Ishan Misra.
\newblock Omnivore: A single model for many visual modalities.
\newblock In \emph{Proceedings of the IEEE/CVF Conference on Computer Vision
  and Pattern Recognition}, pages 16102--16112, 2022.

\bibitem[Goyal et~al.(2017)Goyal, Ebrahimi~Kahou, Michalski, Materzynska,
  Westphal, Kim, Haenel, Fruend, Yianilos, Mueller-Freitag,
  et~al.]{goyal2017something}
Raghav Goyal, Samira Ebrahimi~Kahou, Vincent Michalski, Joanna Materzynska,
  Susanne Westphal, Heuna Kim, Valentin Haenel, Ingo Fruend, Peter Yianilos,
  Moritz Mueller-Freitag, et~al.
\newblock The" something something" video database for learning and evaluating
  visual common sense.
\newblock In \emph{Proceedings of the IEEE International Conference on Computer
  Vision}, pages 5842--5850, 2017.

\bibitem[Hashiguchi and Tamaki(2022)]{hashiguchi2022vision}
Ryota Hashiguchi and Toru Tamaki.
\newblock Vision transformer with cross-attention by temporal shift for
  efficient action recognition.
\newblock \emph{arXiv preprint arXiv:2204.00452}, 2022.

\bibitem[He et~al.(2022)He, Zhou, Ma, Berg-Kirkpatrick, and
  Neubig]{unified_adapter}
Junxian He, Chunting Zhou, Xuezhe Ma, Taylor Berg-Kirkpatrick, and Graham
  Neubig.
\newblock Towards a unified view of parameter-efficient transfer learning.
\newblock 2022.

\bibitem[Houlsby et~al.(2019)Houlsby, Giurgiu, Jastrzebski, Morrone,
  De~Laroussilhe, Gesmundo, Attariyan, and Gelly]{adapter}
Neil Houlsby, Andrei Giurgiu, Stanislaw Jastrzebski, Bruna Morrone, Quentin
  De~Laroussilhe, Andrea Gesmundo, Mona Attariyan, and Sylvain Gelly.
\newblock Parameter-efficient transfer learning for nlp.
\newblock In \emph{ICML}, pages 2790--2799. PMLR, 2019.

\bibitem[Howard et~al.(2017)Howard, Zhu, Chen, Kalenichenko, Wang, Weyand,
  Andreetto, and Adam]{mobilenet}
Andrew~G Howard, Menglong Zhu, Bo~Chen, Dmitry Kalenichenko, Weijun Wang,
  Tobias Weyand, Marco Andreetto, and Hartwig Adam.
\newblock Mobilenets: Efficient convolutional neural networks for mobile vision
  applications.
\newblock \emph{arXiv preprint arXiv:1704.04861}, 2017.

\bibitem[Hu et~al.(2021)Hu, Shen, Wallis, Allen-Zhu, Li, Wang, Wang, and
  Chen]{lora}
Edward~J Hu, Yelong Shen, Phillip Wallis, Zeyuan Allen-Zhu, Yuanzhi Li, Shean
  Wang, Lu~Wang, and Weizhu Chen.
\newblock Lora: Low-rank adaptation of large language models.
\newblock \emph{arXiv preprint arXiv:2106.09685}, 2021.

\bibitem[Jia et~al.(2021)Jia, Yang, Xia, Chen, Parekh, Pham, Le, Sung, Li, and
  Duerig]{jia2021scaling}
Chao Jia, Yinfei Yang, Ye~Xia, Yi-Ting Chen, Zarana Parekh, Hieu Pham, Quoc Le,
  Yun-Hsuan Sung, Zhen Li, and Tom Duerig.
\newblock Scaling up visual and vision-language representation learning with
  noisy text supervision.
\newblock In \emph{International Conference on Machine Learning}, pages
  4904--4916. PMLR, 2021.

\bibitem[Jia et~al.(2022)Jia, Tang, Chen, Cardie, Belongie, Hariharan, and
  Lim]{vpt}
Menglin Jia, Luming Tang, Bor-Chun Chen, Claire Cardie, Serge Belongie, Bharath
  Hariharan, and Ser-Nam Lim.
\newblock Visual prompt tuning.
\newblock \emph{arXiv preprint arXiv:2203.12119}, 2022.

\bibitem[Jiang et~al.(2019)Jiang, Wang, Gan, Wu, and Yan]{jiang2019stm}
Boyuan Jiang, MengMeng Wang, Weihao Gan, Wei Wu, and Junjie Yan.
\newblock Stm: Spatiotemporal and motion encoding for action recognition.
\newblock In \emph{Proceedings of the IEEE/CVF International Conference on
  Computer Vision}, pages 2000--2009, 2019.

\bibitem[Ju et~al.(2021)Ju, Han, Zheng, Zhang, and Xie]{vidprompt}
Chen Ju, Tengda Han, Kunhao Zheng, Ya~Zhang, and Weidi Xie.
\newblock Prompting visual-language models for efficient video understanding.
\newblock \emph{arXiv preprint arXiv:2112.04478}, 2021.

\bibitem[Karpathy et~al.(2014)Karpathy, Toderici, Shetty, Leung, Sukthankar,
  and Fei-Fei]{karpathy2014large}
Andrej Karpathy, George Toderici, Sanketh Shetty, Thomas Leung, Rahul
  Sukthankar, and Li~Fei-Fei.
\newblock Large-scale video classification with convolutional neural networks.
\newblock In \emph{Proceedings of the IEEE conference on Computer Vision and
  Pattern Recognition}, pages 1725--1732, 2014.

\bibitem[Kay et~al.(2017)Kay, Carreira, Simonyan, Zhang, Hillier,
  Vijayanarasimhan, Viola, Green, Back, Natsev, et~al.]{kay2017kinetics}
Will Kay, Joao Carreira, Karen Simonyan, Brian Zhang, Chloe Hillier, Sudheendra
  Vijayanarasimhan, Fabio Viola, Tim Green, Trevor Back, Paul Natsev, et~al.
\newblock The kinetics human action video dataset.
\newblock \emph{arXiv preprint arXiv:1705.06950}, 2017.

\bibitem[Kondratyuk et~al.(2021)Kondratyuk, Yuan, Li, Zhang, Tan, Brown, and
  Gong]{movienet}
D.~Kondratyuk, Liangzhe Yuan, Yandong Li, Li~Zhang, Mingxing Tan, Matthew~A.
  Brown, and Boqing Gong.
\newblock Movinets: Mobile video networks for efficient video recognition.
\newblock \emph{ArXiv}, abs/2103.11511, 2021.

\bibitem[Kuehne et~al.(2011)Kuehne, Jhuang, Garrote, Poggio, and Serre]{hmdb}
Hildegard Kuehne, Hueihan Jhuang, Est{\'\i}baliz Garrote, Tomaso Poggio, and
  Thomas Serre.
\newblock Hmdb: a large video database for human motion recognition.
\newblock In \emph{2011 International conference on computer vision}, pages
  2556--2563. IEEE, 2011.

\bibitem[Kwon et~al.(2020)Kwon, Kim, Kwak, and Cho]{msnet}
Heeseung Kwon, Manjin Kim, Suha Kwak, and Minsu Cho.
\newblock Motionsqueeze: Neural motion feature learning for video
  understanding.
\newblock In \emph{ECCV}, 2020.

\bibitem[Lester et~al.(2021)Lester, Al-Rfou, and
  Constant]{lester-etal-2021-power}
Brian Lester, Rami Al-Rfou, and Noah Constant.
\newblock The power of scale for parameter-efficient prompt tuning.
\newblock In \emph{Proceedings of the 2021 Conference on Empirical Methods in
  Natural Language Processing}, pages 3045--3059, Online and Punta Cana,
  Dominican Republic, November 2021. Association for Computational Linguistics.

\bibitem[Li et~al.(2021{\natexlab{a}})Li, Li, Wang, Wang, and Qiao]{ct_net}
Kunchang Li, Xianhang Li, Yali Wang, Jun Wang, and Y.~Qiao.
\newblock Ct-net: Channel tensorization network for video classification.
\newblock \emph{ArXiv}, abs/2106.01603, 2021{\natexlab{a}}.

\bibitem[Li et~al.(2022)Li, Wang, Zhang, Gao, Song, Liu, Li, and
  Qiao]{li2022uniformer}
Kunchang Li, Yali Wang, Junhao Zhang, Peng Gao, Guanglu Song, Yu~Liu, Hongsheng
  Li, and Yu~Qiao.
\newblock Uniformer: Unifying convolution and self-attention for visual
  recognition.
\newblock \emph{arXiv preprint arXiv:2201.09450}, 2022.

\bibitem[Li and Liang(2021)]{prefix}
Xiang~Lisa Li and Percy Liang.
\newblock Prefix-tuning: Optimizing continuous prompts for generation.
\newblock In \emph{Proceedings of the 59th Annual Meeting of the Association
  for Computational Linguistics and the 11th International Joint Conference on
  Natural Language Processing (Volume 1: Long Papers)}, pages 4582--4597,
  Online, August 2021. Association for Computational Linguistics.

\bibitem[Li et~al.(2021{\natexlab{b}})Li, Wu, Fan, Mangalam, Xiong, Malik, and
  Feichtenhofer]{li2021improved}
Yanghao Li, Chao-Yuan Wu, Haoqi Fan, Karttikeya Mangalam, Bo~Xiong, Jitendra
  Malik, and Christoph Feichtenhofer.
\newblock Improved multiscale vision transformers for classification and
  detection.
\newblock \emph{arXiv preprint arXiv:2112.01526}, 2021{\natexlab{b}}.

\bibitem[Lin et~al.(2019)Lin, Gan, and Han]{tsm}
Ji~Lin, Chuang Gan, and Song Han.
\newblock Tsm: Temporal shift module for efficient video understanding.
\newblock In \emph{Proceedings of the IEEE/CVF International Conference on
  Computer Vision}, pages 7083--7093, 2019.

\bibitem[Liu et~al.(2021{\natexlab{a}})Liu, Ji, Fu, Du, Yang, and
  Tang]{liu2021p}
Xiao Liu, Kaixuan Ji, Yicheng Fu, Zhengxiao Du, Zhilin Yang, and Jie Tang.
\newblock P-tuning v2: Prompt tuning can be comparable to fine-tuning
  universally across scales and tasks.
\newblock \emph{arXiv preprint arXiv:2110.07602}, 2021{\natexlab{a}}.

\bibitem[Liu et~al.(2021{\natexlab{b}})Liu, Lin, Cao, Hu, Wei, Zhang, Lin, and
  Guo]{liu2021swin}
Ze~Liu, Yutong Lin, Yue Cao, Han Hu, Yixuan Wei, Zheng Zhang, Stephen Lin, and
  Baining Guo.
\newblock Swin transformer: Hierarchical vision transformer using shifted
  windows.
\newblock 2021{\natexlab{b}}.

\bibitem[Liu et~al.(2021{\natexlab{c}})Liu, Ning, Cao, Wei, Zhang, Lin, and
  Hu]{liu2021video}
Ze~Liu, Jia Ning, Yue Cao, Yixuan Wei, Zheng Zhang, Stephen Lin, and Han Hu.
\newblock Video swin transformer.
\newblock \emph{arXiv preprint arXiv:2106.13230}, 2021{\natexlab{c}}.

\bibitem[Liu et~al.(2020)Liu, Wang, Wu, Qian, and Lu]{liu2020tam}
Zhaoyang Liu, Limin Wang, Wayne Wu, Chen Qian, and Tong Lu.
\newblock Tam: Temporal adaptive module for video recognition.
\newblock \emph{arXiv preprint arXiv:2005.06803}, 2020.

\bibitem[Luo and Yuille(2019)]{gst}
Chenxu Luo and Alan~L. Yuille.
\newblock Grouped spatial-temporal aggregation for efficient action
  recognition.
\newblock \emph{2019 IEEE International Conference on Computer Vision (ICCV)},
  pages 5511--5520, 2019.

\bibitem[Neimark et~al.(2021)Neimark, Bar, Zohar, and
  Asselmann]{video_transformer}
Daniel Neimark, Omri Bar, Maya Zohar, and Dotan Asselmann.
\newblock Video transformer network.
\newblock \emph{ArXiv}, abs/2102.00719, 2021.

\bibitem[Pan et~al.(2021)Pan, Chen, Shou, Liu, Shao, and Li]{pan2021}
Junting Pan, Siyu Chen, Mike~Zheng Shou, Yu~Liu, Jing Shao, and Hongsheng Li.
\newblock Actor-context-actor relation network for spatio-temporal action
  localization.
\newblock In \emph{Proceedings of the IEEE/CVF Conference on Computer Vision
  and Pattern Recognition (CVPR)}, pages 464--474, June 2021.

\bibitem[Paszke et~al.(2019)Paszke, Gross, Massa, Lerer, Bradbury, Chanan,
  Killeen, Lin, Gimelshein, Antiga, et~al.]{paszke2019pytorch}
Adam Paszke, Sam Gross, Francisco Massa, Adam Lerer, James Bradbury, Gregory
  Chanan, Trevor Killeen, Zeming Lin, Natalia Gimelshein, Luca Antiga, et~al.
\newblock Pytorch: An imperative style, high-performance deep learning library.
\newblock \emph{arXiv preprint arXiv:1912.01703}, 2019.

\bibitem[Patrick et~al.(2021)Patrick, Campbell, Asano, Misra, Metze,
  Feichtenhofer, Vedaldi, and Henriques]{mformer}
Mandela Patrick, Dylan Campbell, Yuki Asano, Ishan Misra, Florian Metze,
  Christoph Feichtenhofer, Andrea Vedaldi, and Jo{\~a}o~F Henriques.
\newblock Keeping your eye on the ball: Trajectory attention in video
  transformers.
\newblock \emph{Advances in Neural Information Processing Systems}, 34, 2021.

\bibitem[Pfeiffer et~al.(2020{\natexlab{a}})Pfeiffer, Kamath, R{\"u}ckl{\'e},
  Cho, and Gurevych]{pfeiffer2020adapterfusion}
Jonas Pfeiffer, Aishwarya Kamath, Andreas R{\"u}ckl{\'e}, Kyunghyun Cho, and
  Iryna Gurevych.
\newblock Adapterfusion: Non-destructive task composition for transfer
  learning.
\newblock \emph{arXiv preprint arXiv:2005.00247}, 2020{\natexlab{a}}.

\bibitem[Pfeiffer et~al.(2020{\natexlab{b}})Pfeiffer, R\"uckl\'{e}, Poth,
  Kamath, Vuli\'{c}, Ruder, Cho, and Gurevych]{pfeiffer2020AdapterHub}
Jonas Pfeiffer, Andreas R\"uckl\'{e}, Clifton Poth, Aishwarya Kamath, Ivan
  Vuli\'{c}, Sebastian Ruder, Kyunghyun Cho, and Iryna Gurevych.
\newblock Adapterhub: A framework for adapting transformers.
\newblock In \emph{Proceedings of the 2020 Conference on Empirical Methods in
  Natural Language Processing (EMNLP 2020): Systems Demonstrations}, pages
  46--54, Online, 2020{\natexlab{b}}. Association for Computational
  Linguistics.

\bibitem[Qin and Eisner(2021)]{qin2021learning}
Guanghui Qin and Jason Eisner.
\newblock Learning how to ask: Querying lms with mixtures of soft prompts.
\newblock \emph{arXiv preprint arXiv:2104.06599}, 2021.

\bibitem[Qiu et~al.(2019)Qiu, Yao, Ngo, Tian, and Mei]{lgd}
Zhaofan Qiu, Ting Yao, C.~Ngo, Xinmei Tian, and Tao Mei.
\newblock Learning spatio-temporal representation with local and global
  diffusion.
\newblock \emph{2019 IEEE/CVF Conference on Computer Vision and Pattern
  Recognition (CVPR)}, pages 12048--12057, 2019.

\bibitem[Radford et~al.(2019)Radford, Wu, Child, Luan, Amodei, Sutskever,
  et~al.]{gpt2}
Alec Radford, Jeffrey Wu, Rewon Child, David Luan, Dario Amodei, Ilya
  Sutskever, et~al.
\newblock Language models are unsupervised multitask learners.
\newblock \emph{OpenAI blog}, 1\penalty0 (8):\penalty0 9, 2019.

\bibitem[Radford et~al.(2021)Radford, Kim, Hallacy, Ramesh, Goh, Agarwal,
  Sastry, Askell, Mishkin, Clark, et~al.]{clip}
Alec Radford, Jong~Wook Kim, Chris Hallacy, Aditya Ramesh, Gabriel Goh,
  Sandhini Agarwal, Girish Sastry, Amanda Askell, Pamela Mishkin, Jack Clark,
  et~al.
\newblock Learning transferable visual models from natural language
  supervision.
\newblock In \emph{International Conference on Machine Learning}, pages
  8748--8763. PMLR, 2021.

\bibitem[Rebuffi et~al.(2017)Rebuffi, Bilen, and Vedaldi]{residualadapter}
Sylvestre-Alvise Rebuffi, Hakan Bilen, and Andrea Vedaldi.
\newblock Learning multiple visual domains with residual adapters.
\newblock 30, 2017.

\bibitem[Rebuffi et~al.(2018)Rebuffi, Bilen, and Vedaldi]{efficientparam}
Sylvestre-Alvise Rebuffi, Hakan Bilen, and Andrea Vedaldi.
\newblock Efficient parametrization of multi-domain deep neural networks.
\newblock pages 8119--8127, 2018.

\bibitem[Ryoo et~al.(2021)Ryoo, Piergiovanni, Arnab, Dehghani, and
  Angelova]{tokenlearner}
Michael Ryoo, AJ~Piergiovanni, Anurag Arnab, Mostafa Dehghani, and Anelia
  Angelova.
\newblock Tokenlearner: Adaptive space-time tokenization for videos.
\newblock \emph{Advances in Neural Information Processing Systems}, 34, 2021.

\bibitem[Sandler et~al.(2018)Sandler, Howard, Zhu, Zhmoginov, and
  Chen]{sandler2018mobilenetv2}
Mark Sandler, Andrew Howard, Menglong Zhu, Andrey Zhmoginov, and Liang-Chieh
  Chen.
\newblock Mobilenetv2: Inverted residuals and linear bottlenecks.
\newblock In \emph{Proceedings of the IEEE conference on computer vision and
  pattern recognition}, pages 4510--4520, 2018.

\bibitem[Sevilla-Lara et~al.(2021)Sevilla-Lara, Zha, Yan, Goswami, Feiszli, and
  Torresani]{sevilla2021only}
Laura Sevilla-Lara, Shengxin Zha, Zhicheng Yan, Vedanuj Goswami, Matt Feiszli,
  and Lorenzo Torresani.
\newblock Only time can tell: Discovering temporal data for temporal modeling.
\newblock In \emph{Proceedings of the IEEE/CVF Winter Conference on
  Applications of Computer Vision}, pages 535--544, 2021.

\bibitem[Sharir et~al.(2021)Sharir, Noy, and Zelnik-Manor]{stam}
Gilad Sharir, Asaf Noy, and Lihi Zelnik-Manor.
\newblock An image is worth 16x16 words, what is a video worth?
\newblock \emph{ArXiv}, abs/2103.13915, 2021.

\bibitem[Shin et~al.(2020)Shin, Razeghi, Logan~IV, Wallace, and
  Singh]{shin2020autoprompt}
Taylor Shin, Yasaman Razeghi, Robert~L Logan~IV, Eric Wallace, and Sameer
  Singh.
\newblock Autoprompt: Eliciting knowledge from language models with
  automatically generated prompts.
\newblock \emph{arXiv preprint arXiv:2010.15980}, 2020.

\bibitem[Singh et~al.(2022)Singh, Gustafson, Adcock, de~Freitas~Reis, Gedik,
  Kosaraju, Mahajan, Girshick, Doll{\'a}r, and van~der Maaten]{swag}
Mannat Singh, Laura Gustafson, Aaron Adcock, Vinicius de~Freitas~Reis, Bugra
  Gedik, Raj~Prateek Kosaraju, Dhruv Mahajan, Ross Girshick, Piotr Doll{\'a}r,
  and Laurens van~der Maaten.
\newblock Revisiting weakly supervised pre-training of visual perception
  models.
\newblock In \emph{Proceedings of the IEEE/CVF Conference on Computer Vision
  and Pattern Recognition}, pages 804--814, 2022.

\bibitem[Soomro et~al.(2012)Soomro, Zamir, and Shah]{ucf101}
Khurram Soomro, Amir~Roshan Zamir, and Mubarak Shah.
\newblock Ucf101: A dataset of 101 human actions classes from videos in the
  wild.
\newblock \emph{arXiv preprint arXiv:1212.0402}, 2012.

\bibitem[Stickland and Murray(2019)]{stickland2019bert}
Asa~Cooper Stickland and Iain Murray.
\newblock Bert and pals: Projected attention layers for efficient adaptation in
  multi-task learning.
\newblock In \emph{International Conference on Machine Learning}, pages
  5986--5995. PMLR, 2019.

\bibitem[Sun et~al.(2019)Sun, Myers, Vondrick, Murphy, and
  Schmid]{sun2019videobert}
Chen Sun, Austin Myers, Carl Vondrick, Kevin Murphy, and Cordelia Schmid.
\newblock Videobert: A joint model for video and language representation
  learning.
\newblock In \emph{Proceedings of the IEEE/CVF International Conference on
  Computer Vision}, pages 7464--7473, 2019.

\bibitem[Sung et~al.(2021)Sung, Cho, and Bansal]{vladapter}
Yi-Lin Sung, Jaemin Cho, and Mohit Bansal.
\newblock Vl-adapter: Parameter-efficient transfer learning for
  vision-and-language tasks.
\newblock \emph{arXiv preprint arXiv:2112.06825}, 2021.

\bibitem[Tian et~al.(2020)Tian, Krishnan, and Isola]{tian2020contrastive}
Yonglong Tian, Dilip Krishnan, and Phillip Isola.
\newblock Contrastive multiview coding.
\newblock In \emph{European conference on computer vision}, pages 776--794.
  Springer, 2020.

\bibitem[Tong et~al.(2022)Tong, Song, Wang, and Wang]{tong2022videomae}
Zhan Tong, Yibing Song, Jue Wang, and Limin Wang.
\newblock Videomae: Masked autoencoders are data-efficient learners for
  self-supervised video pre-training.
\newblock \emph{arXiv preprint arXiv:2203.12602}, 2022.

\bibitem[Tran et~al.(2015)Tran, Bourdev, Fergus, Torresani, and
  Paluri]{tran2015learning}
Du~Tran, Lubomir Bourdev, Rob Fergus, Lorenzo Torresani, and Manohar Paluri.
\newblock Learning spatiotemporal features with 3d convolutional networks.
\newblock In \emph{Proceedings of the IEEE international conference on computer
  vision}, pages 4489--4497, 2015.

\bibitem[Tran et~al.(2018{\natexlab{a}})Tran, Wang, Torresani, Ray, LeCun, and
  Paluri]{r21}
Du~Tran, Heng Wang, Lorenzo Torresani, Jamie Ray, Yann LeCun, and Manohar
  Paluri.
\newblock A closer look at spatiotemporal convolutions for action recognition.
\newblock In \emph{Proceedings of the IEEE conference on Computer Vision and
  Pattern Recognition}, pages 6450--6459, 2018{\natexlab{a}}.

\bibitem[Tran et~al.(2018{\natexlab{b}})Tran, Wang, Torresani, Ray, LeCun, and
  Paluri]{tran2018closer}
Du~Tran, Heng Wang, Lorenzo Torresani, Jamie Ray, Yann LeCun, and Manohar
  Paluri.
\newblock A closer look at spatiotemporal convolutions for action recognition.
\newblock In \emph{Proceedings of the IEEE conference on Computer Vision and
  Pattern Recognition}, pages 6450--6459, 2018{\natexlab{b}}.

\bibitem[Tran et~al.(2019)Tran, Wang, Torresani, and Feiszli]{csn}
Du~Tran, Heng Wang, L.~Torresani, and Matt Feiszli.
\newblock Video classification with channel-separated convolutional networks.
\newblock \emph{2019 IEEE/CVF International Conference on Computer Vision
  (ICCV)}, pages 5551--5560, 2019.

\bibitem[Wang et~al.(2020{\natexlab{a}})Wang, Tran, Torresani, and
  Feiszli]{corrnet}
Heng Wang, Du~Tran, L.~Torresani, and Matt Feiszli.
\newblock Video modeling with correlation networks.
\newblock \emph{2020 IEEE/CVF Conference on Computer Vision and Pattern
  Recognition (CVPR)}, pages 349--358, 2020{\natexlab{a}}.

\bibitem[Wang et~al.(2018{\natexlab{a}})Wang, Xiong, Wang, Qiao, Lin, Tang, and
  Van~Gool]{wang2018temporal}
Limin Wang, Yuanjun Xiong, Zhe Wang, Yu~Qiao, Dahua Lin, Xiaoou Tang, and Luc
  Van~Gool.
\newblock Temporal segment networks for action recognition in videos.
\newblock \emph{IEEE transactions on pattern analysis and machine
  intelligence}, 41\penalty0 (11):\penalty0 2740--2755, 2018{\natexlab{a}}.

\bibitem[Wang et~al.(2020{\natexlab{b}})Wang, Tong, Ji, and Wu]{tdn}
Limin Wang, Zhan Tong, Bin Ji, and Gangshan Wu.
\newblock Tdn: Temporal difference networks for efficient action recognition.
\newblock \emph{ArXiv}, abs/2012.10071, 2020{\natexlab{b}}.

\bibitem[Wang et~al.(2021)Wang, Xing, and Liu]{wang2021actionclip}
Mengmeng Wang, Jiazheng Xing, and Yong Liu.
\newblock Actionclip: A new paradigm for video action recognition.
\newblock \emph{arXiv preprint arXiv:2109.08472}, 2021.

\bibitem[Wang et~al.(2018{\natexlab{b}})Wang, Girshick, Gupta, and
  He]{wang2018non}
Xiaolong Wang, Ross Girshick, Abhinav Gupta, and Kaiming He.
\newblock Non-local neural networks.
\newblock In \emph{Proceedings of the IEEE conference on computer vision and
  pattern recognition}, pages 7794--7803, 2018{\natexlab{b}}.

\bibitem[Wei et~al.(2021)Wei, Fan, Xie, Wu, Yuille, and
  Feichtenhofer]{wei2021masked}
Chen Wei, Haoqi Fan, Saining Xie, Chao-Yuan Wu, Alan Yuille, and Christoph
  Feichtenhofer.
\newblock Masked feature prediction for self-supervised visual pre-training.
\newblock \emph{arXiv preprint arXiv:2112.09133}, 2021.

\bibitem[Xie et~al.(2018)Xie, Sun, Huang, Tu, and Murphy]{s3d}
Saining Xie, Chen Sun, Jonathan Huang, Zhuowen Tu, and Kevin Murphy.
\newblock Rethinking spatiotemporal feature learning: Speed-accuracy trade-offs
  in video classification.
\newblock In \emph{Proceedings of the European conference on computer vision
  (ECCV)}, pages 305--321, 2018.

\bibitem[Xu et~al.(2021{\natexlab{a}})Xu, Ghosh, Huang, Arora, Aminzadeh,
  Feichtenhofer, Metze, and Zettlemoyer]{xu2021vlm}
Hu~Xu, Gargi Ghosh, Po-Yao Huang, Prahal Arora, Masoumeh Aminzadeh, Christoph
  Feichtenhofer, Florian Metze, and Luke Zettlemoyer.
\newblock Vlm: Task-agnostic video-language model pre-training for video
  understanding.
\newblock \emph{arXiv preprint arXiv:2105.09996}, 2021{\natexlab{a}}.

\bibitem[Xu et~al.(2021{\natexlab{b}})Xu, Ghosh, Huang, Okhonko, Aghajanyan,
  Metze, Zettlemoyer, and Feichtenhofer]{xu2021videoclip}
Hu~Xu, Gargi Ghosh, Po-Yao Huang, Dmytro Okhonko, Armen Aghajanyan, Florian
  Metze, Luke Zettlemoyer, and Christoph Feichtenhofer.
\newblock Videoclip: Contrastive pre-training for zero-shot video-text
  understanding.
\newblock In \emph{EMNLP}, 2021{\natexlab{b}}.

\bibitem[Yan et~al.(2022)Yan, Xiong, Arnab, Lu, Zhang, Sun, and Schmid]{mtv}
Shen Yan, Xuehan Xiong, Anurag Arnab, Zhichao Lu, Mi~Zhang, Chen Sun, and
  Cordelia Schmid.
\newblock Multiview transformers for video recognition.
\newblock In \emph{Proceedings of the IEEE/CVF Conference on Computer Vision
  and Pattern Recognition}, pages 3333--3343, 2022.

\bibitem[Yu et~al.(2022)Yu, Wang, Vasudevan, Yeung, Seyedhosseini, and
  Wu]{yu2022coca}
Jiahui Yu, Zirui Wang, Vijay Vasudevan, Legg Yeung, Mojtaba Seyedhosseini, and
  Yonghui Wu.
\newblock Coca: Contrastive captioners are image-text foundation models.
\newblock \emph{arXiv preprint arXiv:2205.01917}, 2022.

\bibitem[Zhang et~al.(2021)Zhang, Hao, and Ngo]{tokenshift}
Hao Zhang, Yanbin Hao, and Chong-Wah Ngo.
\newblock Token shift transformer for video classification.
\newblock In \emph{Proceedings of the 29th ACM International Conference on
  Multimedia}, pages 917--925, 2021.

\bibitem[Zhang et~al.(2020)Zhang, Sax, Zamir, Guibas, and Malik]{sidetune}
Jeffrey~O Zhang, Alexander Sax, Amir Zamir, Leonidas Guibas, and Jitendra
  Malik.
\newblock Side-tuning: a baseline for network adaptation via additive side
  networks.
\newblock pages 698--714. Springer, 2020.

\bibitem[Zhang et~al.(2022)Zhang, Zhou, and Liu]{zhang2022neural}
Yuanhan Zhang, Kaiyang Zhou, and Ziwei Liu.
\newblock Neural prompt search.
\newblock \emph{arXiv preprint arXiv:2206.04673}, 2022.

\bibitem[Zhou et~al.(2021)Zhou, Yang, Loy, and Liu]{coop}
Kaiyang Zhou, Jingkang Yang, Chen~Change Loy, and Ziwei Liu.
\newblock Learning to prompt for vision-language models.
\newblock \emph{arXiv preprint arXiv:2109.01134}, 2021.

\bibitem[Zhou et~al.(2022)Zhou, Yang, Loy, and Liu]{cocoop}
Kaiyang Zhou, Jingkang Yang, Chen~Change Loy, and Ziwei Liu.
\newblock Conditional prompt learning for vision-language models.
\newblock In \emph{CVPR}, 2022.

\bibitem[Zhu et~al.(2020)Zhu, Tran, Sevilla-Lara, Yang, Feiszli, and
  Wang]{faster}
Linchao Zhu, Du~Tran, Laura Sevilla-Lara, Yi~Yang, Matt Feiszli, and Heng Wang.
\newblock Faster recurrent networks for efficient video classification.
\newblock In \emph{Proceedings of the AAAI Conference on Artificial
  Intelligence}, volume~34, pages 13098--13105, 2020.

\bibitem[Zolfaghari et~al.(2018)Zolfaghari, Singh, and Brox]{eco}
Mohammadreza Zolfaghari, Kamaljeet Singh, and Thomas Brox.
\newblock Eco: Efficient convolutional network for online video understanding.
\newblock In \emph{Proceedings of the European conference on computer vision
  (ECCV)}, pages 695--712, 2018.

\end{thebibliography}
